\documentclass[letterpaper]{article}

\usepackage{booktabs}
\usepackage{aaai18}  

\usepackage{times}  
\usepackage{helvet}  
\usepackage{courier}  
\usepackage{url}  
\usepackage{graphicx}  
\usepackage{multirow}
\usepackage{subfigure}
\usepackage{enumitem}
\usepackage{amsmath}
\usepackage{amssymb}
\usepackage{bm}
\usepackage{multicol}
\usepackage{chngcntr}
\usepackage{pifont}

\usepackage{gensymb}
\usepackage[usenames, dvipsnames]{xcolor}

\usepackage{breqn}
\usepackage{mathptmx} 

\usepackage{adjustbox}
\usepackage{array}

\newcolumntype{R}[2]{%
    >{\adjustbox{angle=#1,lap=\width-(#2)}\bgroup}%
    l%
    <{\egroup}%
}
\newcommand*\rot{\multicolumn{1}{R{45}{1em}}}

\newcommand{\inline}[1]{\citeauthor{#1} \shortcite{#1}}

\newcommand{\cmark}{\ding{51}}%
\newcommand{\xmark}{\ding{55}}%

\begin{document}
%
\title{Measuring Catastrophic Forgetting in Neural Networks}
\author{Ronald Kemker, Marc McClure, Angelina Abitino, Tyler Hayes, Christopher Kanan\\
Chester F. Carlson Center for Imaging Science\\Rochester Institute of Technology\\54 Lomb Memorial Drive\\Rochester NY 14623\\
{\tt\small \{rmk6217, mcm5756, tlh6792, kanan\}@rit.edu , aabitin1@swarthmore.edu}}
\nocopyright

\maketitle
\begin{abstract}
  	Deep neural networks are used in many state-of-the-art systems for machine perception. Once a network is trained to do a specific task, e.g., bird classification, it cannot easily be trained to do new tasks, e.g., incrementally learning to recognize additional bird species or learning an entirely different task such as flower recognition.  When new tasks are added, typical deep neural networks are prone to catastrophically forgetting previous tasks. Networks that are capable of assimilating new information incrementally, much like how humans form new memories over time, will be more efficient than re-training the model from scratch each time a new task needs to be learned.  There have been multiple attempts to develop schemes that mitigate catastrophic forgetting, but these methods have not been directly compared, the tests used to evaluate them vary considerably, and these methods have only been evaluated on small-scale problems (e.g., MNIST). In this paper, we introduce new metrics and benchmarks for directly comparing five different mechanisms designed to mitigate catastrophic forgetting in neural networks: regularization, ensembling, rehearsal, dual-memory, and sparse-coding.  Our experiments on real-world images and sounds show that the mechanism(s) that are critical for optimal performance vary based on the incremental training paradigm and type of data being used, but they all demonstrate that the catastrophic forgetting problem has yet to be solved. 
\end{abstract}

\section{Introduction}

\noindent While the basic architecture and training algorithms behind deep neural networks (DNNs) are over 30 years old, interest in them has never been greater in both industry and the artificial intelligence research community. Owing to far larger datasets, increases in computational power, and innovations in activation functions, DNNs have achieved near-human or super-human abilities on a number of problems, including image classification \cite{he2016deep}, speech-to-text~\cite{khilari2015review}, and face identification~\cite{schroff2015facenet}. These algorithms power most of the recent advances in semantic segmentation ~\cite{long2015fully}, visual question answering~\cite{kafle2017visual}, and reinforcement learning~\cite{mnih-atari-2013}. While these systems have become more capable, the standard multi-layer perceptron (MLP) architecture and typical training algorithms cannot handle incrementally learning new tasks or categories without catastrophically forgetting previously learned training data. Fixing this problem is critical to making agents that incrementally improve after deployment. For non-embedded or personalized systems, catastrophic forgetting is often overcome simply by storing new training examples and then re-training either the entire network from scratch or possibly only the last few layers. In both cases, retraining uses \emph{both} the previously learned examples and the new examples, randomly shuffling them so that they are independent and identically distributed (iid). Retraining can be slow, especially if a dataset has millions or billions of instances.

\begin{figure}[t]
    \centering
    \includegraphics[width=0.99\linewidth]{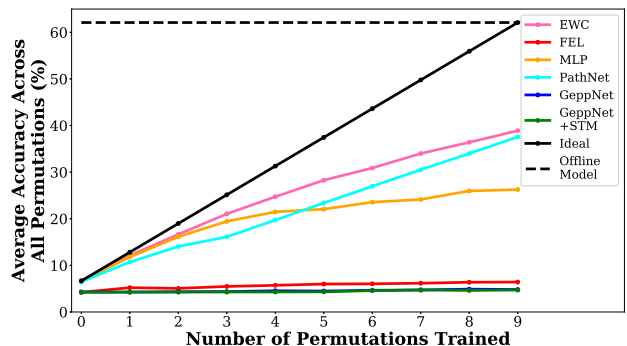}
    
    \caption{Catastrophic forgetting impairs incremental learning in neural networks.  As a network is incrementally trained (solid lines), ideally its performance would match that of a model trained offline with all of the data upfront (dashed line). In this paper, we develop methods and benchmarks for measuring catastrophic forgetting. Our experiments show that even methods designed to prevent catastrophic forgetting perform significantly worse than an offline model.  Incremental learning is key to many real-world applications because it allows the model to adapt after being deployed.}
    \label{fig:up_front}
\end{figure}

Catastrophic forgetting was first recognized in MLPs almost 30 years ago \cite{mccloskey1989catastrophic}. Since then, there have been multiple attempts to mitigate this phenomenon \cite{hinton1987using,robins1995catastrophic,goodrich2014unsupervised,draelos2016neurogenesis,ren2017life,fernando2017pathnet,kirkpatrick2017overcoming}. However, these methods vary considerably in how they train and evaluate their models and they focus on small datasets, e.g., MNIST. It is not clear if these methods will scale to larger datasets containing hundreds of categories. In this paper, we remedy this problem by providing a comprehensive empirical review of methods to mitigate catastrophic forgetting across a variety of new metrics. While catastrophic forgetting occurs in unsupervised frameworks \cite{draelos2016neurogenesis,goodrich2014unsupervised,triki2017encoder}, we focus on supervised classification. \textbf{Our major contributions are:}
\begin{itemize}[noitemsep,nolistsep]
\item We demonstrate that despite popular claims~\cite{kirkpatrick2017overcoming}, catastrophic forgetting is not solved.
\item We establish new benchmarks with novel metrics for measuring catastrophic forgetting. Previous work has focused on MNIST, which contains low-resolution images and only 10 classes. Instead, we use real-world image/audio classification datasets containing 100-200 classes.  We show that, although existing models perform well on MNIST for a variety of different incremental learning problems, performance drops significantly with more challenging datasets. 
\item We identified five common mechanisms for mitigating catastrophic forgetting: 1) regularization, 2) ensembling, 3) rehearsal, 4) dual-memory models, and 5) sparse-coding.  Unlike previous work, we directly compare these distinct approaches. 
\end{itemize}

\subsection{Problem Formulation}
In this paper, we study catastrophic forgetting in MLP-based neural networks that are incrementally trained for classification tasks. In our setup, the labeled training dataset $D$ is organized into $T$ study sessions (batches), i.e., $D = \left\{ {B_t } \right\}_{t = 1}^T$. Each study session $B_t$ consists of $N_t$ labeled training data points, i.e., $
B_t  = \left\{ {\left( {{\mathbf{x}}_j ,y_j } \right)} \right\}_{j = 1}^{N_t } $, where $\mathbf{x}_j \in \mathbb{R}^d$ and $y_j$ is a discrete label. $N_t$ is variable across sessions. The model is only permitted to learn sessions sequentially, in order. At time $t$ the network can only learn from study session $B_t$; however, models may use auxiliary memory to store previously observed sessions, but this memory use must be reported. We do not assume sessions are iid, e.g., some sessions may contain data from only a single category. In between  sessions, the model may be evaluated on test data. Because this paper's focus is catastrophic forgetting, we focus less on representation learning and obtain feature vectors using embeddings from pre-trained networks. Note that in some other papers, new sessions are called new `tasks.' We refer to the first study session as the model's `base set knowledge.'

\section{Why Does Catastrophic Forgetting Occur?}

Catastrophic forgetting in neural networks occurs because of the stability-plasticity dilemma~\cite{abraham2005memory}. The model requires sufficient plasticity to acquire new tasks, but large weight changes will cause forgetting by disrupting previously learned representations. Keeping the network's weights stable prevents previously learned tasks from being forgotten, but too much stability prevents the model from learning new tasks. Prior research has tried to solve this problem using two broad approaches. The first is to try to keep new and old representations separate, which can be done using distributed models, regularization, and ensembling. The second is to prevent the forgetting of prior knowledge simply by training on the old tasks (or some facsimile of them) as well as new tasks, thereby preventing the old tasks from being forgotten. Besides requiring costly re-learning of previous examples and additional storage, this scheme is still not as effective as simply combining the new data with the old data and completely re-training the model from scratch. This solution is inefficient as it prevents the development of deployable systems that are capable of learning new tasks over the course of their lifetime.

\section{Previous Surveys}

 \inline{french1999catastrophic} exhaustively reviewed mechanisms for preventing catastrophic forgetting that were explored in the 1980s and 1990s. \inline{goodfellow2013empirical} compared different activation functions and learning algorithms to see how they affected
catastrophic forgetting, but these methods were not explicitly designed to mitigate catastrophic forgetting. The authors concluded that the learning algorithms have a larger impact, which is what we focus on in our paper. They sequentially trained a network on two separate tasks using three different scenarios: 1) identical tasks with different forms of input, 2) similar tasks, and 3) dissimilar tasks.  We adopt a similar paradigm, but our experiments involve a much larger number of tasks.  We also focus on methods explicitly designed to mitigate catastrophic forgetting.

\inline{soltoggio2017born} reviewed neural networks that can adapt their plasticity over time, which they called Evolved Plastic Artificial Neural Networks.  Their review covered a wide-range of brain-inspired algorithms and also identified that the field lacks appropriate benchmarks. However, they did not conduct any experiments or establish benchmarks for measuring catastrophic forgetting. We remedy this gap in the literature by establishing large-scale benchmarks for evaluating catastrophic forgetting in neural networks, and we compare methods that use five distinct mechanisms for mitigating it.

\section{Mitigating Catastrophic Forgetting}
While not exhaustive, we have identified five main approaches that have been pursued for mitigating catastrophic forgetting in MLP-like architectures, which we describe in the next subsections. We describe the models we have selected in greater detail in the Experimental Setup section.

\subsection{Regularization Methods}

Regularization methods add constraints to the network's weight updates, so that a new session is learned without interfering with prior memories.  \inline{hinton1987using} implemented a network that had both `fast' and `slow' training weights.  The fast weights had high plasticity and were easily affected by changes to the network, and the `slow' weights had high stability and were harder to adapt. This kind of dual-weight architecture is similar in idea to dual-network models, but has not been proven to be sufficiently powerful to learn a large number of new tasks. Elastic weight consolidation (EWC) \cite{kirkpatrick2017overcoming} adds a constraint to the loss function that directs plasticity away from weights that contribute the most to previous tasks.   We use EWC to evaluate the regularization mechanism.

\subsection{Ensemble Methods}
Ensemble methods attempt to mitigate catastrophic forgetting either by explicitly or implicitly training multiple classifiers together and then combining them to generate the final prediction. For the explicit methods, such as Learn++ and TradaBoost, this prevents forgetting because an entirely new sub-network is trained for a new session~\cite{polikar2001learn++,dai2007boosting}. However, memory usage will scale with the number of sessions, which is highly non-desirable. Moreover, this prevents portions of the network from being re-used for the new session. Two methods that try to alleviate the memory usage problem are Accuracy Weighted Ensembles and Life-long Machine Learning~\cite{wang2003mining,ren2017life}. These methods automatically decide whether a sub-network should be removed or added to the ensemble. 

PathNet can be considered as an implicit ensemble method~\cite{fernando2017pathnet}. It uses a genetic algorithm to find an optimal path through a fixed-size neural network for each study session. The weights in this path are then frozen; so that when new sessions are learned, the knowledge is not lost. In contrast to the explicit ensembles, the base network's size is fixed and it is possible for learned representations to be re-used which allows for smaller, more deployable models.  The authors showed that PathNet learned subsequent tasks more quickly, but not how well earlier tasks were retained.  We have selected PathNet to evaluate the ensembling mechanism, and we show how well it retains pre-trained information.  

\subsection{Rehearsal Methods}
Rehearsal methods try to mitigate catastrophic forgetting by mixing data from earlier sessions with the current session being learned~\cite{robins1995catastrophic}. The cost is that this requires storing past data, which is not resource efficient.  Pseudorehearsal methods use the network to generate pseudopatterns \cite{robins1995catastrophic} that are combined with the session currently being learned. Pseudopatterns allow the network to stabilize older memories without the requirement for storing all previously observed training data points. \inline{draelos2016neurogenesis} used this approach to incrementally train an autoencoder, where each session contained images from a specific category.  After the autoencoder learned a particular session, they passed the session's data through the encoder and stored the output statistics.  During replay, they used these statistics and the decoder network to generate the appropriate pseudopatterns for each class. 

The GeppNet model proposed by \inline{Gepperth2016} reserves its training data to replay after each new class was trained.  This model used a self-organizing map (SOM) as a hidden-layer to topologically reorganize the data from the input layer (i.e., clustering the input onto a 2-D lattice). We use this model to explore the value of rehearsal. 

\subsection{Dual-Memory Models}
Dual-memory models are inspired by memory consolidation in the mammalian brain, which is thought to store memories in two distinct neural networks. Newly formed memories are stored in a brain region known as the hippocampus. These memories are then slowly transferred/consolidated to the pre-frontal cortex during sleep. Several algorithms based on these ideas have been created. 
Early work used fast (hippocampal) and slow (cortical) training networks to separate pattern-processing areas, and they passed pseudopatterns back and forth to consolidate recent and remote memories \cite{french1997pseudo}. In general, dual-memory models incorporate rehearsal, but not all rehearsal-based models are dual-memory models.

Another model proposed by \inline{Gepperth2016}, which we denote GeppNet+STM, stores new inputs that yield a highly uncertain prediction into a short-term memory (STM) buffer. This model then seeks to consolidate the new memories into the entire network during a separate sleep phase. They showed that GeppNet+STM could incrementally learn MNIST classes without forgetting previously trained ones. We use GeppNet and GeppNet+STM to evaluate the dual-memory approach.

\subsection{Sparse-Coding Methods}

Catastrophic forgetting occurs when new internal representations interfere with previously learned ones~\cite{french1999catastrophic}. Sparse representations can reduce the chance of this interference; however, sparsity can impair generalization and ability to learn new tasks~\cite{sharkey1995analysis}.

Two models that implicitly use sparsity are CALM and ALCOVE. To learn new data, CALM searches among competing nodes to see which nodes have not been committed to another representation \cite{murre2014learning}.  ALCOVE is a shallow neural network that uses a sparse distance-based representation, which allows the weights assigned to older tasks to be largely unchanged when the network is presented with new data \cite{kruschke1992alcove}. The Sparse Distributed Memory (SDM) is a convolution-correlation model that uses sparsity to reduce the overlap between internal representations~\cite{kanerva1988sparse}.  CHARM and TODAM are also convolution-correlation models that use internal codings to ensure that new input representations remain orthogonal to one another~\cite{murdock1983distributed,eich1982composite}.

The Fixed Expansion Layer (FEL) model creates sparse representations by fixing the network's weights and specifying neuron triggering conditions~\cite{FEL}. FEL uses excitatory and inhibitory fixed weights to sparsify the input, which gates the weight updates throughout the network.  This enables the network to retain prior learned mappings and reduce representational overlap.  We use FEL to evaluate the sparsity mechanism.

\section{Experimental Setup}

We explore how well methods to mitigate catastrophic forgetting scale on hard datasets involving fine-grained image and audio classification.  These datasets were chosen because they contain 1) different data modalities (image and audio), 2) a large number of classes, and 3) a small number of samples per class.  These datasets are more meaningful (real-world problems) and more practical than MNIST. We also use MNIST to showcase the value of these real-world datasets.  See Table \ref{table:data} for dataset statistics.

\subsection{Dataset Description}

\subsubsection{MNIST} MNIST is a classic dataset in machine learning containing 10 digit classes. Its grayscale images are $28 \times 28$. 

\subsubsection{CUB-200}
Caltech-UCSD Birds-200 (CUB-200) is an image classification dataset containing 200 different bird species~\cite{WahCUB_200_2011}.  We use the 2011 version. Each high-resolution image is turned into a 2048-dimensional vector with ResNet-50 \cite{he2016deep}, which is a deep convolutional neural network (DCNN) pre-trained on ImageNet \cite{ILSVRC15}.  Extracting image features from the last hidden (fully-connected) layer of pre-trained {DCNNs} is a common practice in computer vision. We report mean-per-class accuracy, which is the CUB-200 standard.  

\begin{table}[t!]
\centering \footnotesize
\begin{tabular}{@{}rccc@{}}\toprule
  & \textbf{MNIST}&\textbf{CUB-200} & \textbf{AudioSet}\\ \midrule
 \textbf{Classification Task }& Gray Image  & RGB Image  & Audio \\
 \textbf{Classes}  & 10  & 200 & 100 \\
 \textbf{Feature Shape} & 784  & 2,048 & 1,280 \\
 \textbf{Train Samples}& 50,000 & 5,994 & 28,779\\
 \textbf{Test Samples} & 10,000 & 5,794 & 5,523\\
  \textbf{Train Samples/Class}  & 5,421-6,742 & 29-30  & 250-300  \\
  \textbf{Test Samples/Class}	& 892-1,135	  & 11-30  & 43-62  \\
\bottomrule 
\end{tabular}
\caption{Dataset Specifications}
\label{table:data}
\end{table}

\subsubsection{AudioSet}
AudioSet~\cite{gemmeke2017audio} is a hierarchically organized audio classification dataset built from YouTube videos.  It has over 2 million human-labeled, 10 second sound bytes drawn from one or more of 632 classes.  We used the pre-extracted frame-wise features from AudioSet concatenated in order. These features were extracted using a variant ResNet-50 for audio data \cite{hershey2017CNN}, which was pre-trained on an early version of the YouTube-8m dataset \cite{DBLP:journals/corr/Abu-El-HaijaKLN16}. We used 100 classes from AudioSet, none of which were super or sub-classes of each other. The classes did not have any restrictions based on the AudioSet ontology, and all of them had a quality estimation of over 70\%.  Each audio sample can have multiple labels, so we chose training and testing samples that were labeled with only 1 of the 100 classes. 

\subsection{Models Evaluated}

We evaluated five models that correspond to each of the five mechanisms described in the previous section: 1) EWC, 2) PathNet, 3) GeppNet, 4) GeppNet+STM, and 5)  FEL. To choose the number of parameters to use across models, we established a baseline MLP architecture that performed well for CUB-200 and AudioSet when trained offline.  The goal is to determine which mechanism(s) work best for various incremental learning paradigms. To provide a fair comparison, the number of parameters in each model were chosen to be as close as possible to the number of parameters in the baseline MLP.  We optimized each model's  hyperparameters to work well for our benchmarks, which are given in supplemental materials~\footnote{Supplemental materials provided at the end of our arXiv submission: \url{https://arxiv.org/abs/1708.02072}}.  The supplemental materials provides the stopping criteria for each model as defined by their creators, which involved 1) training for a fixed period of time or 2) using test accuracy to stop training early.

\subsubsection{Standard Multi-Layer Perceptron}
\label{sss:mlp}
For our baseline, we use a standard MLP.  Its architecture was chosen by optimizing performance using the entire training set for both CUB-200 and AudioSet, i.e., it was trained offline. The offline MLP achieves 62.1\%  accuracy on the CUB-200 test set and 46.1\% on the AudioSet test set.  We did a hyperparameter search for the number of units per hidden layer (32-4,096), number of hidden layers (2-3), and weight decay parameter (0, $10^{-4}$, $5\cdot 10^{-4}$).  The MLP model was also trained incrementally to measure the severity of catastrophic forgetting.  

\subsubsection{Elastic Weight Consolidation}
\label{sss:ewc}
EWC adds an additional constraint to the loss function $L\left(\theta\right)$, i.e.,
\begin{equation}
\label{eq:ewc}
\footnotesize
L\left(\theta\right) = L_t\left(\theta\right) + \sum\limits_{i} \frac{\lambda}{2} F_i\left(\theta_i - \theta_{A,i}^*\right)^2,
\end{equation}
where $L\left(\theta\right)$ is the combined loss function, $\theta$ is the network's parameters, $L_t\left(\theta\right)$ is the loss for session $B_t$, $\lambda$ is a hyperparameter that indicates how important the old task(s) are compared to the new task, $F$ is the Fisher information matrix, and $\theta_A^*$ are the  trainable parameters (weights and biases) important to previously trained tasks.  The Fisher matrix is used to constrain the weights important to previously learned tasks to their original value; that is, plasticity is directed to the trainable parameters that contribute the least to performing previously trained tasks.  The size of the hidden-layer was chosen to match the baseline MLP's capacity. 

\subsubsection{PathNet}
\label{sss:pathnet}
PathNet is a fixed size neural network that uses a genetic algorithm to find the optimal path through the network. Only this path is trainable when learning a particular session, which is why the authors described their model as an evolutionary dropout network. PathNet creates an independent output layer for each task in order to preserve previously trained tasks, and it cannot be used without modifications for incremental class learning. Since entire portions of the network are sequentially frozen as new tasks are learned, there is a risk of PathNet losing its ability to learn once the maximum capacity is reached. PathNet's capacity was chosen to match the capacity of the MLP baseline.  

\subsubsection{GeppNet}
\label{sss:som}
GeppNet and GeppNet+STM are biologically-inspired approaches that use rehearsal to mitigate forgetting. In these models, training the initial task starts by initializing the SOM-layer, which is used to project the probability density of the input to a higher two-dimensional lattice.  The SOM-layer features are passed to a linear regression classification layer to make a prediction.  During training, the SOM-layer is initialized with the first session for a fixed-period of time, and then the SOM- and classification-layers are trained jointly. The SOM-layer is only updated when a training example is determined by the model to be novel (i.e., using the prediction probabilities to generate a confidence measure).  After GeppNet has been trained on the initial session for a fixed period of time, it incrementally learns subsequent sessions.  
GeppNet performs updates to the SOM-layer and classification-layer when a training example is considered novel. When GeppNet+STM detects novelty, it instead uses a fixed-size short-term memory buffer to store that training example, which then replays it during a sleep phase. The sleep phase repeats after a fixed number of training iterations. Since the replay queue has a fixed-size (i.e., older examples are replaced), the GeppNet+STM model will train more efficiently than GeppNet.  GeppNet stores all previous training data and replays it along with the previous data during a portion of its incremental learning step.  GeppNet+STM also stores all previous and new training data; however, each training example is only replayed if the model is uncertain on the prediction. In addition, GeppNet+STM  is capable of making real-time predictions by determining if the desired memory is in short-term memory (the memory buffer) or in long-term storage (the SOM- and classification-layers). 

\subsubsection{Fixed Expansion Layer}

FEL uses sparsity to mitigate catastrophic forgetting~\cite{FEL}. FEL is a two hidden-layer MLP where the second hidden-layer (FEL-layer) has a higher capacity than the first fully-connected layer, but the weights are sparse and remain fixed through training.  Each FEL-layer unit is only connected to a subset of the units in the first hidden layer, and these connections are split between excitatory and inhibitory weights.  Only a subset of the FEL-layer units are allowed to have non-zero output to the final classification layer, which causes only some of the units in the first hidden layer to be updated. 

\section{Experiments and Results}

We have established three benchmark experiments for measuring catastrophic forgetting: 
\begin{enumerate}
\item \textbf{Data Permutation Experiment} - The elements of every feature vector are randomly permuted, with the permutation  held constant within a session, but varying across sessions. The model is evaluated on its ability to recall data learned in prior study sessions.  Each session contains the same number of examples. 
\item \textbf{Incremental Class Learning} - After learning the base set, each new session learned contains only a single class. 
\item \textbf{Multi-Modal Learning} - The model incrementally learns different datasets, e.g., learn image classification and then audio classification. 
\end{enumerate}

For the data permutation and incremental class learning experiments, each model was also evaluated on MNIST.  The goal is to examine whether results on MNIST generalize to the real-world datasets. More  results, including detailed plots, can be found in the supplementary materials. 

\subsection{Evaluation Metrics}

We propose three new metrics to evaluate a model's ability to retain prior sessions while still learning new knowledge, 

\begin{equation}
\label{eq:base}
\footnotesize
\Omega_{base} = \frac{1}{T-1} \sum_{i=2}^T \frac{\alpha_{base,i}}{\alpha_{ideal}} 
\end{equation} 
\begin{equation}
\small
\label{eq:new}
\Omega_{new} = \frac{1}{T-1} \sum_{i=2}^T \alpha_{new,i}   
\end{equation}

\begin{equation}
\label{eq:combined}
\small
\Omega_{all} = \frac{1}{T-1} \sum_{i=2}^T \frac{\alpha_{all,i}}{\alpha_{ideal}} %
\end{equation}
where  $T$ is the total number of sessions, $\alpha_{new,i}$ is the test accuracy for session $i$ immediately after it is learned,  $\alpha_{base,i}$ is the test accuracy on the first session (base set) after $i$ new sessions have been learned, $\alpha_{all,i}$ is the test accuracy of all of the test data for the classes seen to this point, and $\alpha_{ideal}$ is the offline MLP accuracy on the base set, which we assume is the ideal performance. $\Omega_{base}$ and $\Omega_{new}$ are normalized area under the curve metrics.  $\Omega_{base}$ measures a model's retention of the first session, after learning in later study sessions. $\Omega_{new}$ measures the model's ability to immediately recall new tasks. By normalizing $\Omega_{base}$ and $\Omega_{all}$  by $\alpha_{ideal}$, the results will be easier to compare between datasets. Unless a model exceeds $\alpha_{ideal}$, results will be between $\left[0,1\right]$, which enables comparison between datasets.  $\Omega_{all}$ computes how well a model both retains prior knowledge and acquires new information. 

\subsection{Experimental Results}
\subsubsection{Data Permutation Experiment}
This experiment evaluates a model's ability to retain multiple representations of the dataset, with each representation learned sequentially. These representations are created by randomly permuting the elements of the input feature vectors, with the random permutation changing between sessions. An identically permuted test set is used along with each session. This paradigm provides overlapping tasks in which each session contains the same information and categories,  so each session is of equal complexity. This paradigm is identical to that used by \inline{goodfellow2013empirical}  and \inline{kirkpatrick2017overcoming}.

\begin{table*}[t!]
\centering \footnotesize
\begin{tabular}{@{}cc|ccc|ccc|ccc|c|c@{}}\toprule
 \multirow{2}{*}{\textbf{Model}} & \multirow{2}{*}{\textbf{Dataset}}  & \multicolumn{3}{c|}{\textbf{Data Permutation}} & \multicolumn{3}{c|}{\textbf{Incremental Class}}& \multicolumn{3}{c|}{\textbf{Multi-Modal}} & \textbf{Memory} & \textbf{Model}  \\
 & & $\Omega_{base}$ & $\Omega_{new}$ & $\Omega_{all}$& $\Omega_{base}$ & $\Omega_{new}$ & $\Omega_{all}$  &  $\Omega_{base}$ & $\Omega_{new}$ & $\Omega_{all}$ & \textbf{Constraints} & \textbf{Size (MB)} \\
\midrule
 \multirow{3}{*}{\textbf{MLP}}	  	&\textbf{MNIST} & 0.434 & 0.996 & 0.702 &   0.060 & 1.000 & 0.181  & N/A   & N/A & N/A & \multirow{3}{*}{Fixed-size} & 1.91\\
 								  	&\textbf{CUB}   & 0.488 & 0.917 & 0.635 &   0.020 & 1.000 & 0.031  & 0.327 & 0.412 & 0.610 &  & 4.24 \\
 				                  	&\textbf{AS}    & 0.186 & 0.957 & 0.446 &   0.016 & 1.000 & 0.044  & 0.197 & 0.609 & 0.589 &  & 2.85\\[1ex]
 \multirow{3}{*}{\textbf{EWC}}	  	&\textbf{MNIST} & 0.437 & 0.992 & 0.746 &   0.001 & 1.000 & 0.133  & N/A   & N/A   & N/A   & \multirow{3}{*}{Fixed-size}& 3.83\\
 								  	&\textbf{CUB}   & 0.765 & 0.869 & 0.762 &   0.105 & 0.000 & 0.094  & 0.944 & 0.369 & 0.872 &  & 8.48\\
 				                  	&\textbf{AS}    & 0.129 & 0.687 & 0.251 &   0.021 & 0.580 & 0.034  & 1.000 & 0.588 & 0.984 &  & 5.70\\[1ex]
 \multirow{3}{*}{\textbf{PathNet}}	&\textbf{MNIST} & 0.687 & 0.887 & 0.848 &   N/A   & N/A   & N/A      & N/A   & N/A   & N/A   &  New output & 2.80\\
 								  	&\textbf{CUB}   & 0.538 & 0.701 & 0.655 &   N/A   & N/A   & N/A      & 0.908 & 0.376 & 0.862 & layer for & 7.46\\
 				                  	&\textbf{AS}    & 0.414 & 0.750 & 0.615 &   N/A   & N/A   & N/A      & 0.069 & 0.540 & 0.469 & each task & 4.68\\[1ex]
 \multirow{3}{*}{\textbf{GeppNet}}  &\textbf{MNIST} & 0.912 & 0.242 & 0.364 &   0.960 & 0.824 & 0.922  & N/A   & N/A   & N/A   & Stores all & 190.08\\
 								  	&\textbf{CUB}   & 0.606 & 0.029 & 0.145 &   0.628 & 0.640 & 0.585  & 0.156 & 0.010 & 0.089 & training & 53.48 \\
 				                  	&\textbf{AS}   	& 0.897 & 0.170 & 0.343 &   0.984 & 0.458 & 0.947  & 0.913 & 0.005 & 0.461 & data & 150.38\\[1ex]
 \multirow{3}{*}{\textbf{GeppNet+STM}}&\textbf{MNIST}& 0.892 & 0.212 & 0.326 &  0.919 & 0.599 & 0.824  & N/A   & N/A   & N/A   & Stores all & 191.02\\
								  	&\textbf{CUB}   & 0.615 & 0.020 & 0.142 &   0.727 & 0.232 & 0.626  & 0.031 & 0.329 & 0.026 & training & 55.94 \\
							     	&\textbf{AS}    & 0.820 & 0.041 & 0.219 &   1.007 & 0.355 & 0.920  & 0.829 & 0.005 & 0.418 & data & 151.92\\[1ex]
 \multirow{3}{*}{\textbf{FEL}}	  	&\textbf{MNIST} & 0.117 & 0.990 & 0.279 &   0.451 & 1.000 & 0.439  & N/A   & N/A   & N/A   &  \multirow{3}{*}{Fixed-size}& 4.54\\
 								  	&\textbf{CUB}   & 0.043 & 0.764 & 0.184 &   0.316 & 1.000 & 0.361  & 0.110 & 0.329 & 0.412 & & 6.16 \\
 				                  	&\textbf{AS}    & 0.081 & 0.848 & 0.239 &   0.283 & 1.000 & 0.240  & 0.473 & 0.320 & 0.494 & & 6.06 \\[1ex]
\bottomrule 
\end{tabular}
\caption{Results on MNIST, CUB-200 (CUB), and AudioSet (AS) for our evaluation metrics as well as model size (in MB) for each model/dataset combination.}
\label{table:results}
\end{table*}

Results are given in Table \ref{table:results}. In nearly every case, $\Omega_{all}$ is greater for MNIST than on CUB-200 or AudioSet, demonstrating the need for alternative incremental learning benchmarks.  To some extent, EWC, PathNet, GeppNet, and GeppNet+STM retain prior knowledge without forgetting; however, GeppNet and GeppNet+STM fail to learn new sessions.  PathNet and EWC seem to retain base knowledge while still learning new information; however, PathNet performs better on AudioSet and MNIST, while EWC performs better on CUB-200 (see discussion).

\subsubsection{Incremental Class Learning}

In the incremental class learning experiment, a model's ability to sequentially learn new classes is tested.  The first session learned contains training data from half of the classes in each dataset: 5 for MNIST, 100 for CUB-200, and 50 for AudioSet.  Once this base set was learned, each subsequent session contained training data from a single new class. We measure mean-per-class accuracy on the base set after each new class is learned to assess a model's long-term memory. We also calculate the accuracy of each class after it is trained to ensure the model is still learning, and we calculate the performance of all previously learned classes.

PathNet is incapable of learning new classes incrementally because it creates a separate output layer for each additional session. Accessing that output layer during prediction time requires a priori information on which session the model needs to access. This means PathNet requires the testing label to make the appropriate prediction, which would result in a misleading high test accuracy. For this reason, we omitted PathNet from this experiment.

\begin{figure*}[t!]
  \centering
  \subfigure[MNIST]{%
  \includegraphics[width=0.32\linewidth]{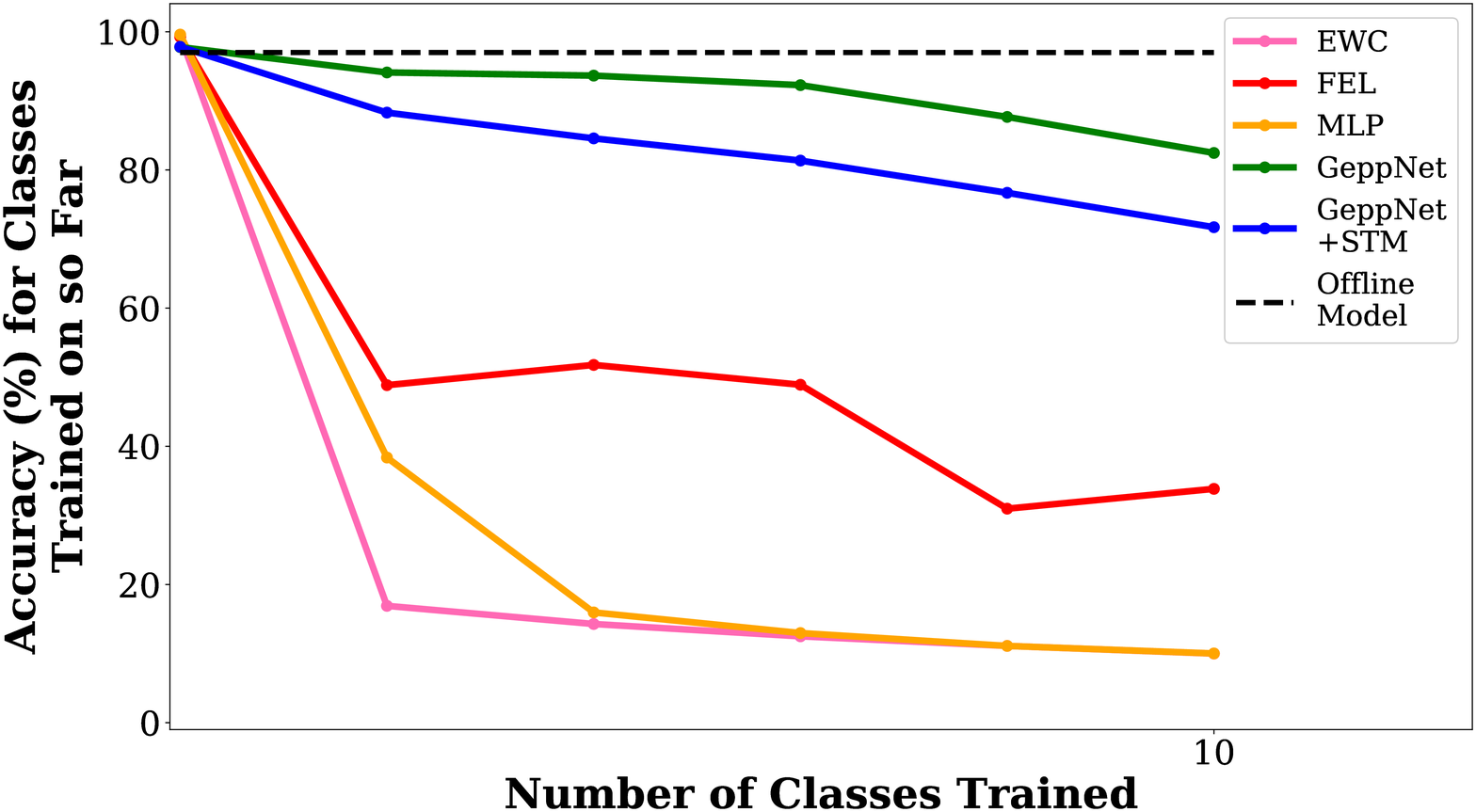}
  \label{fig:result_a}} 
  \subfigure[CUB-200]{%
  \includegraphics[width=0.32\linewidth]{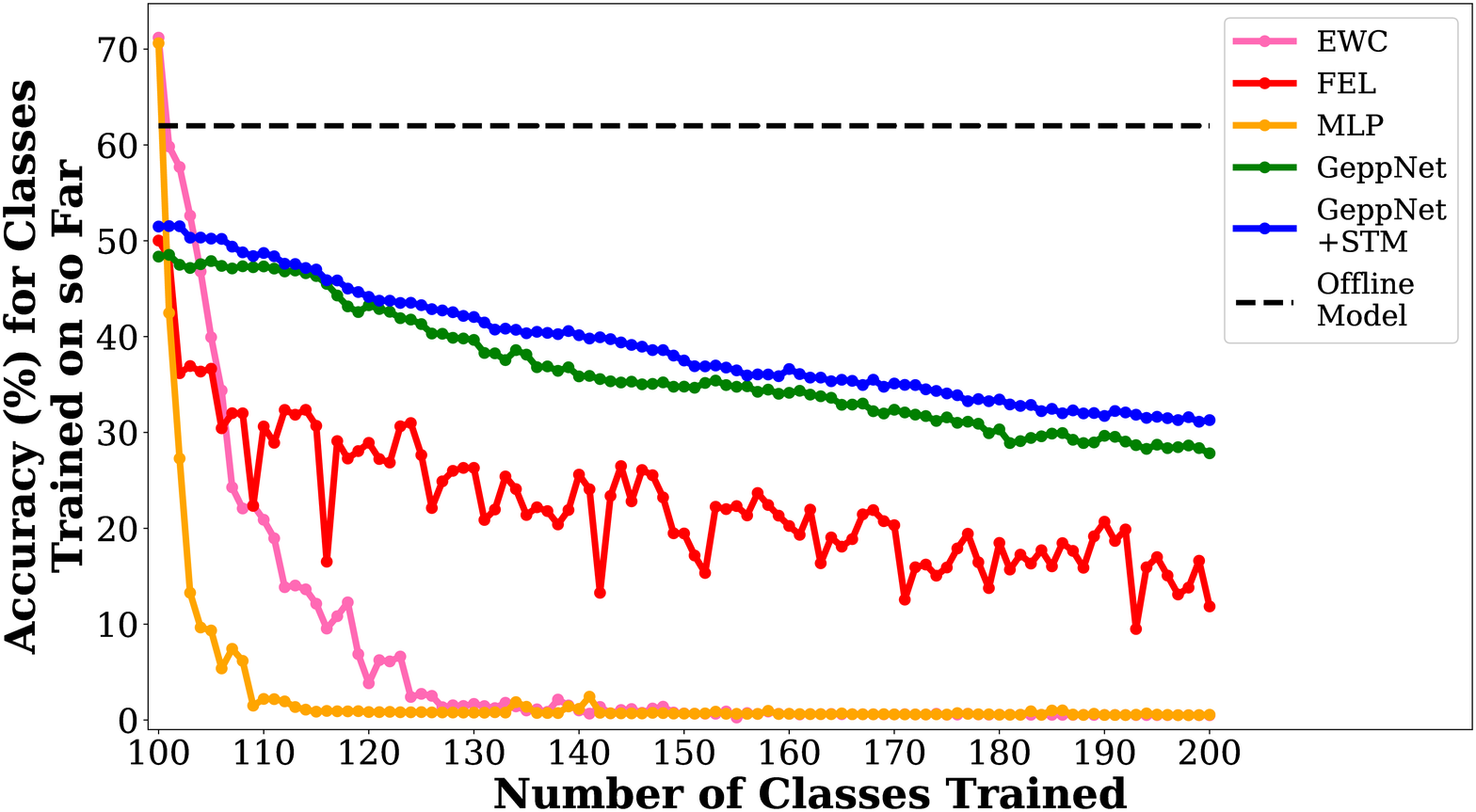}
  \label{fig:result_b}}  
  \subfigure[AudioSet]{%
  \includegraphics[width=0.32\linewidth]{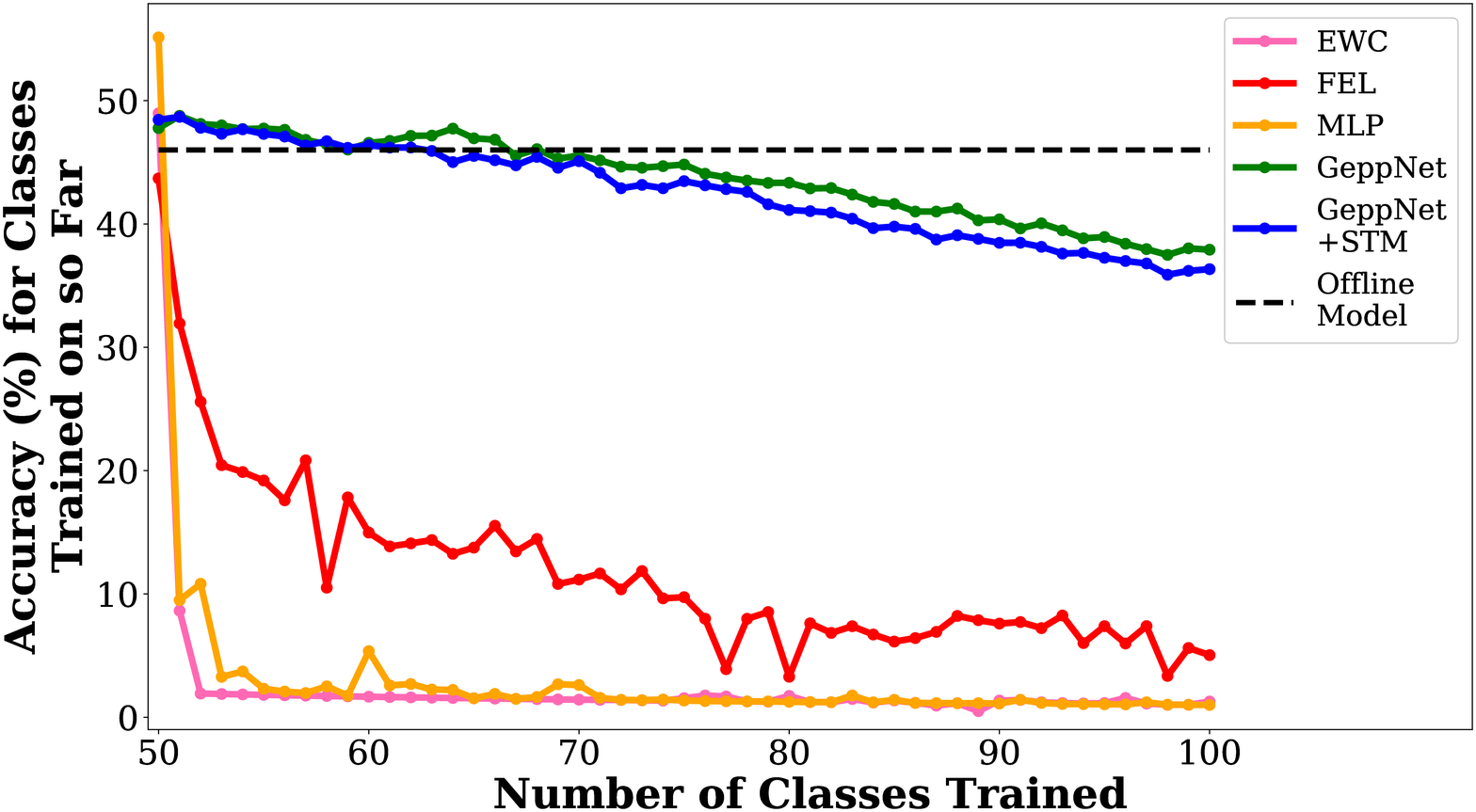}
  \label{fig:result_c}}
  \caption{{Results for incremental class learning experiment.}  This shows the mean-class test accuracy for all classes seen so far.}
  \label{fig:results}
\end{figure*}

Results are summarized in Table \ref{table:results} and Fig. \ref{fig:results} contains plots for the mean-per-class test accuracy for all classes seen so far. The only models that were able to both retain the base knowledge and learn new classes were GeppNet, GeppNet+STM, and FEL, with the clear winner being GeppNet.  Much like the data permutation experiment, the CUB-200 and AudioSet results were noticeably lower than the MNIST results.  GeppNet+STM did well at retaining the base set, but it struggled to learn new classes on CUB-200 and AudioSet.  This could be because the model only trains during sleep for efficiency reasons.  Additionally, the short-term memory buffer is emptied after training each study session, which is when the model is evaluated.  This type of model could work better in a real-time environment.  FEL  learned new classes well, but suffered from forgetting of the base set.  FEL may benefit from  larger model capacity, but this would require more memory/processing power.

\subsubsection{Multi-Modal Experiment}
The goal of the multi-modal experiment is to determine if a network can learn and retain multiple dissimilar tasks that have 1) inputs with different dimensionality and feature distributions and 2) a different number of classes.  A system like this could be useful for learning tasks that have multi-modal data using a single network and could be more efficient than building a separate neural network for each modality (e.g., video has visual and audio information).  In this experiment, we evaluated each model's ability to perform image and audio classification with CUB-200 and AudioSet respectively. In this experiment, there are only two incrementally learned sessions, where each session contains AudioSet or CUB-200. We compare learning AudioSet first then CUB-200 (AS/CUB) and learning CUB then AudioSet (CUB/AS).

The ResNet features obtained from CUB-200 have a higher dimensionality than the AudioSet features, so we zero-padded the AudioSet input to match the dimensionality of CUB-200.  This experiment is done by training one dataset to completion followed by training the other dataset to completion (and vice-versa). Once both modalities have been trained, we evaluate the first modality that was trained in order to measure how well the model was able to retain what it learned about the first task. 

Table \ref{table:results} shows summary results for the multi-modal experiment, where the corresponding row is the modality that was trained first, i.e. the row for CUB-200 is where CUB-200 is learned first followed by AudioSet.  Additional results are in supplementary materials. Although several models perform well at one of the two experiments, EWC is the only model capable of preserving the first modality while also learning the second modality for both cases, which we explore further in the discussion.  

\section{Discussion}

In our paper we introduced new metrics and benchmarks for measuring catastrophic forgetting. Our results reveal that none of the methods we tested solve catastrophic forgetting, while also enabling the learning of new information. Table \ref{table:summary} summarizes these results for each of our experiments by averaging $\Omega_{all}$ over all datasets. While no method excels at incremental learning, some perform better than others.

PathNet performed best overall on the data permutation experiments, with the exception of CUB-200. However, PathNet requires being told which session each test instance is from, whereas the other models do not use this information. This may give it an unfair advantage. PathNet works by locking the optimal path for a given session. Because permuting the data does not reduce feature overlap, the model requires more trainable weights (less feature sharing) to build a discriminative model, causing PathNet to saturate (freeze all weights) more quickly.  When PathNet reaches the saturation point, the only trainable parameters are in the output layer. While EWC was the second best performing method in the permutation experiments, it only redirects plasticity instead of freezing trainable weights.

\begin{table}[t!]
\centering \footnotesize
\begin{tabular}{@{}cccc@{}}\toprule
 \multirow{2}{*}{\textbf{Model}}& \textbf{Data}  & \textbf{Incremental} & \multirow{2}{*}{\textbf{Multi-Modal}} \\
 & \textbf{Permutation} & \textbf{Class} &   \\
\midrule
 \textbf{MLP}	 		&0.594 &0.085 & 0.600 \\
 \textbf{EWC}	 		&0.586 &0.087 &\textbf{0.913}  \\
 \textbf{PathNet} 	 	&\textbf{0.706} &N/A &0.666  \\
 \textbf{GeppNet}	 	&0.284 &\textbf{0.818} &0.275  \\
 \textbf{GeppNet+STM}	&0.229 &0.790 &0.222  \\
 \textbf{FEL}	 		&0.234 &0.347 &0.453  \\
\bottomrule 
\end{tabular}
\caption{{Summary of Experimental Results. Average of $\Omega_{all}$ over MNIST, CUB-200, and AudioSet results.}}
\label{table:summary}
\end{table}

Both GeppNet variants performed best at incremental class learning.  These models make slow, gradual changes to the network that are inspired by memory consolidation during sleep. For these models, the SOM-layer was fixed to $23\times23$ to have the same number of trainable parameters as the other models.  With 100-200 classes, this corresponds to 2-5 hidden layer neurons per class respectively.  The experiments on MNIST in \inline{Gepperth2016} used 90 hidden-layer neurons per class, so their performance may improve if their model capacity was significantly increased, but this would demand more memory and computation. 

EWC performed best on the multi-modal experiment. This may be because features between the two modalities are non-redundant.  We hypothesize that EWC is a better choice for separating non-redundant data and PathNet may work well when working with data that has different, but not entirely dissimilar, representations. To explore this, we used the Fast Correlation Based Filter proposed by \inline{yu2003feature} to show the features in MNIST and AudioSet are more redundant than those in CUB-200 (see supplemental material). The performance of EWC and PathNet for both the data permutation and multi-modal experiments are consistent with this hypothesis.

Table \ref{table:results} shows the memory constraints and usage of each model. While we kept the number of trainable parameters roughly the same across all models in their hidden layers, some require additional memory resources. PathNet generates a new output layer for each session. Both GeppNet variants store all training data and rehearse over it during their incremental learning stage.  The creators of EWC stored validation data from all previous sessions and used it to minimize forgetting when learning a new session. This was not done in our experiments to fairly compare it to the other models, which only had access to validation data for the current session. 

Table \ref{table:time} shows the total time to train each model for the data permutation and incremental class learning experiments using CUB-200.  Both variants of GeppNet are orders of magnitude slower because they train the model one sample at a time.  PathNet is also very slow at the data permutation task because the optimal path through a large DCNN needs to be found for each permutation.  The fixed-size models are noticeably faster; however, only EWC was effective at mitigating catastrophic forgetting (in the data permutation and multi-modal experiments). 


\begin{table}[t!]
\centering \footnotesize
\begin{tabular}{@{}ccc@{}}\toprule
 \multirow{2}{*}{\textbf{Model}}& \textbf{Data}  & \textbf{Incremental}  \\
 & \textbf{Permutation} & \textbf{Class}   \\
\midrule
 \textbf{MLP}	 & 16& 15  \\
 \textbf{EWC}	 & 16& 13  \\
\textbf{PathNet} &1,385 & N/A   \\
 \textbf{GeppNet} &507 &1,123   \\
 \textbf{GeppNet+STM} & 179& 410   \\
 \textbf{FEL}	 & 53 & 8 \\
\bottomrule 
\end{tabular}
\caption{Training time (minutes) for each model on CUB-200.}
\label{table:time}
\end{table}

In general, models that expand as a function of the number of sessions and those that are allowed to store data from prior sessions may have limited real-world application. In our opinion, methods for mitigating catastrophic forgetting should have the amount of total memory they use constrained. While our summary statistics did not take this into account, it is an important factor in deploying a method that learns incrementally. This is the reason we chose to keep the number of trainable parameters fixed across all models. 

An alternative would have been to tune the number of trainable parameters in each model for each experiment, which is what we did for the data permutation and incremental class learning experiments  as well (see Supplemental Materials for details).  Although in most cases the base performance increased, there were no changes to any of our conclusions on which model/mechanism yielded superior performance.  The one interesting thing we did observe is that the sparsity model (i.e. FEL) can sometimes improve significantly; however, the cost is a 40x increase in the model’s memory footprint.  This reinforces our claim that a model that only uses the sparsity mechanism to mitigate catastrophic forgetting may not be ideal in a deployed environment.  We urge future incremental learning algorithm creators to take memory footprint into account, especially when comparing to other models. 

Our metrics could be expanded to other training paradigms such as reinforcement learning, unsupervised learning, etc.  In reinforcement learning, the agent learns an initial study-session (e.g. an ATARI game), which represents the base knowledge.  We would track the performance of the base-knowledge as the model learns additional games and ensure that the model is learning new games as well.  The main difference is that the performance metrics would be normalized by the maximum performance for each study-session when the model only has to learn that single study session.  In unsupervised learning, we could follow the experiments performed by \cite{draelos2016neurogenesis} where the metrics would be the same, but we would train the models using a different loss function (e.g. reconstruction error).

\begin{table}[t!]
\centering \footnotesize
\begin{tabular}{@{}ccccccc@{}}\toprule
\textbf{Model}& \rot{\textbf{Incremental Class}} & \rot{\textbf{Similar Data}} & \rot{\textbf{Dissimilar Data}} & \rot{\textbf{Memory Efficient} } & \rot{\textbf{Trains Quickly}} & \hspace{10 mm}   \\
\midrule
 \textbf{MLP}	 &\xmark &\xmark &\xmark & \cmark & \cmark  \\
 \textbf{EWC}	 &\xmark &\xmark &\cmark& \cmark&\cmark \\
\textbf{PathNet} &\xmark & \cmark &\xmark  &\xmark &\xmark  \\
 \textbf{GeppNet} & \cmark &\xmark &\xmark&\xmark&\xmark   \\
 \textbf{GeppNet+STM} & \cmark &\xmark &\xmark&\xmark&\xmark  \\
 \textbf{FEL}	 & \xmark & \xmark& \xmark& \xmark& \cmark \\
\bottomrule 
\end{tabular}
\caption{Summary of the optimal performer on the incremental class learning, data permutation (Similar Data), and multi-modal (Dissimilar Data) experiments, as well as the memory/computational efficiency of each model.}
\label{table:end}
\end{table}

\section{Conclusion}
In this paper, we developed new metrics for evaluating catastrophic forgetting. We identified five families of mechanisms for mitigating catastrophic forgetting in DNNs. We found that performance on MNIST was significantly better than on the larger datasets we used.  Using our new metrics, experimental results (summarized in Table \ref{table:end}) show that 1) a combination of rehearsal/pseudo-rehearsal and dual-memory systems are optimal for learning new classes incrementally, and  2) regularization and ensembling are best at separating multiple dissimilar sessions in a common DNN framework.  Although the rehearsal system performed reasonably well, it required retaining all training data for replay.  This type of system may not be scalable for a real-world lifelong learning system; however, it does indicate that models that use pseudorehearsal could be a viable option for real-time incremental learning systems.  Future work on lifelong learning frameworks should involve combinations of these mechanisms.  While some models perform better than others in different scenarios, our work shows that catastrophic forgetting is not solved by any single method.  This is because there is no model that is capable of assimilating new information while simultaneously and efficiently preserving the old. We urge the community to use larger datasets in future work.

\section{Acknowledgements}
A. Abitino was supported by NSF Research Experiences for Undergraduates (REU) award \#1359361 to R. Dube. We thank NVIDIA for the generous donation of a Titan X GPU.

\newpage 
\bibliography{bibtex} 
\bibliographystyle{aaai}

\newpage
\appendix
\counterwithin{figure}{section}
\counterwithin{table}{section}
\renewcommand{\thefigure}{S\arabic{figure}}
\renewcommand{\thetable}{S\arabic{table}}
\setcounter{figure}{0} 
\setcounter{table}{0} 

\section{Supplemental Material}

\subsection{Training Parameters}

\begin{table}[h!]
\centering \footnotesize
\begin{tabular}{@{}ll@{}}\toprule
 \textbf{Training Parameter} & \textbf{Value} \\
 \midrule
 Units per Layer         & 400 \\
 Hidden Layers           & 2 \\
 Mini-Batch Size              & 256 \\
 Hidden-Layer Activation & ReLU \\
 Optimizer               & Nadam  \\
 Initial Learning Rate   & 8e-4 \\
 Convergence Criteria    & Early-Stopping \\
\bottomrule \\
\end{tabular}
\caption{Training parameters for the MLP baseline model}
\label{table:mlp_specs}
\end{table}

\begin{table}[h!]
\centering \footnotesize
\begin{tabular}{@{}ll@{}}\toprule
 \textbf{Training Parameter} & \textbf{Value} \\
 \midrule
 Units per Layer         & 400 \\
 Hidden Layers           & 2 \\
 Mini-Batch Size              & 250 \\
 Hidden-Layer Activation & ReLU \\
 Optimizer               & Adam  \\
 Initial Learning Rate   & 2e-4 \\
 Convergence Criteria    & Early Stopping \\
\bottomrule \\

\end{tabular}
\caption{Training parameters for the EWC model}
\label{table:ewc_specs}
\end{table}

\begin{table}[h!]
\centering \footnotesize
\begin{tabular}{@{}ll@{}}\toprule
 \textbf{Training Parameter} & \textbf{Value} \\
 \midrule
 Layers (L)              & 2 \\
 Modules (M)             & 10 \\
 Modules per Layer (N)   & 5 \\
 Units per Module		 & 80 \\
 Mini-Batch Size              & 16 \\
 Hidden Layer Activation & ReLU \\
 Optimizer               & Adam  \\
 Initial Learning Rate   & 2e-3 \\
 Convergence Criteria    & Early Stopping \\
\bottomrule \\
\end{tabular}
\caption{Training parameters for the PathNet model.}
\label{table:pathnet_specs}
\end{table}

\begin{table}[h!]
\centering \footnotesize
\begin{tabular}{@{}ll@{}}\toprule
 \textbf{Training Parameter} & \textbf{Value} \\
 \midrule
SOM Shape & 23x23  \\
 Mini-Batch Size              & 1 \\
 Hidden-Layer Activation & Custom \\
 Initial SOM Learning Rate   & 0.1 \\
 MLP Learning Rate       & 1e-3\\
 Modulation Threshold ($\theta_{m}^{inc}$) & 0.5 \\
 \multirow{2}{*}{Convergence Criteria} & 80,000 iterations (Base) \\
 									   & 20,000 iterations (Incremental) \\
\bottomrule \\
\end{tabular}
\caption{Training parameters for the GeppNet and GeppNet+STM models.}
\label{table:som_specs}
\end{table}

\begin{table}[h!]
\centering \footnotesize
\begin{tabular}{@{}ll@{}}\toprule
 \textbf{Training Parameter} & \textbf{Value} \\
 \midrule
 Hidden Layer Units                 & 400 \\
 \multirow{3}{*}{FEL Layer Neurons} & 1200 (CUB-200) \\
                                    & 2000 (AudioSet) \\
                                    & 1200 (Multi-modal experiment)\\
 Mini-Batch Size                         & 256 \\
 Optimizer               			& Adam  \\
 Initial Learning Rate   			& 2e-2 \\
 Convergence Criteria    			& Early Stopping \\
\bottomrule \\
\end{tabular}
\caption{Training parameters for the FEL model.}
\label{table:fel_specs}
\end{table}

\subsection{Reproduction Validation Experiments}

In the following section we have documented our results from our reproduction of some of the experiments previously done with each of these models. PathNet is not included in this section because we were able to get the model directly from \inline{fernando2017pathnet}.

\begin{figure}[th!]
    \centering
    \includegraphics[width=0.9\linewidth]{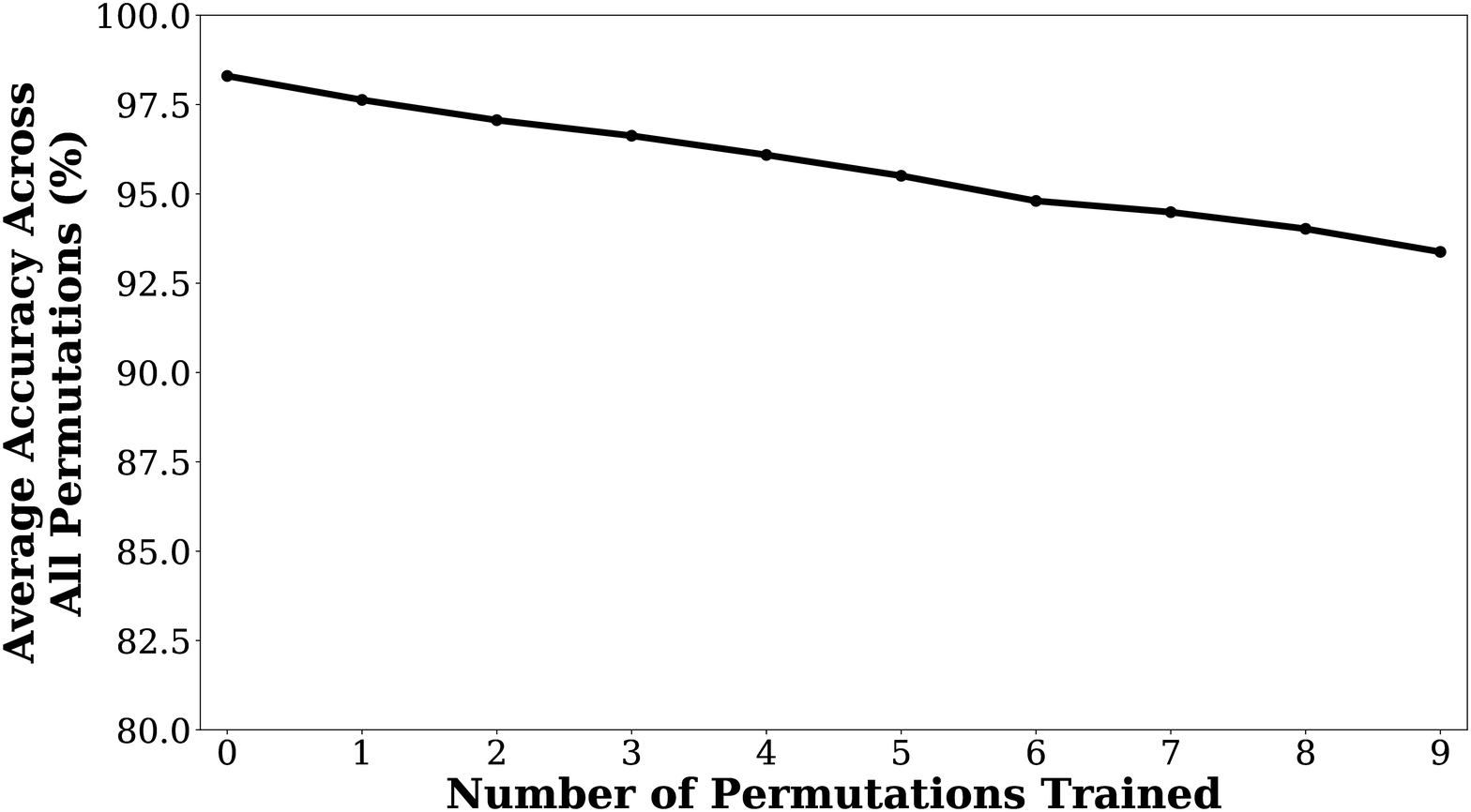}
    \caption{Our results for EWC on the MNIST experiment created by \inline{kirkpatrick2017overcoming}.}
    \label{fig:verify-ewc}
\end{figure}

Fig. \ref{fig:verify-ewc} demonstrates the results for our implementation of EWC from the MNIST experiment proposed by \inline{kirkpatrick2017overcoming}.  Unlike the training methodology employed in our main paper, for the reproduction, we used the validation data from previous permutations to help retain previously trained tasks which is consistent with the original implementation of EWC.  The results show the mean test accuracy across all permuted datasets seen so far.  We performed a grid search across the hyperparameters (hidden layer size and learning rate) listed in the paper.  Our model performs similarly to the one in \inline{kirkpatrick2017overcoming}.

Table \ref{table:som_verify} contains results from our GeppNet and GeppNet+STM model verification experiment.  Each test ``Inc-X''  involves training the base with every class except for ``X'' and then adding Class ``X'' incrementally.  \inline{Gepperth2016} do not list specific percentages for each test, but the results in Table \ref{table:som_verify} are similar to the author's.  The three reported metrics, in order, include the accuracy of the base prior to the incremental training step, the accuracy of the new class after the incremental training step, and the overall accuracy of all test data after the incremental training step.

\begin{table}[ht!]
\centering \footnotesize
\begin{tabular}{@{}rcccc@{}}\toprule
\multirow{2}{*}{\textbf{Test}} & \multirow{2}{*}{\textbf{Model}} & \multicolumn{3}{c}{\textbf{Accuracy}} \\
& &\textbf{Base} & \textbf{New Class} & \textbf{Overall}\\
\midrule
\multirow{2}{*}{\textbf{Inc-0}} & \textbf{GeppNet}     & 92.9 & 93.2 & 92.6 \\
                                & \textbf{GeppNet+STM} & 92.4 & 83.2 & 90.4 \\[1ex] 
\multirow{2}{*}{\textbf{Inc-1}} & \textbf{GeppNet}     & 92.9 & 97.8 & 93.3 \\
                                & \textbf{GeppNet+STM} & 93.1 & 97.0 & 93.1 \\[1ex] 
\multirow{2}{*}{\textbf{Inc-2}} & \textbf{GeppNet}     & 93.3 & 81.8 & 92.0 \\
                                & \textbf{GeppNet+STM} & 92.9 & 84.0 & 90.4 \\[1ex] 
\multirow{2}{*}{\textbf{Inc-3}} & \textbf{GeppNet}     & 93.9 & 80.6 & 91.9 \\
                                & \textbf{GeppNet+STM} & 94.1 & 55.6 & 88.8 \\[1ex] 
\multirow{2}{*}{\textbf{Inc-4}} & \textbf{GeppNet}     & 94.1 & 72.0 & 90.6 \\
                                & \textbf{GeppNet+STM} & 93.6 & 94.5 & 85.3 \\[1ex] 
\multirow{2}{*}{\textbf{Inc-5}} & \textbf{GeppNet}     & 94.1 & 74.0 & 91.5 \\
                                & \textbf{GeppNet+STM} & 93.9 & 62.0 & 89.8 \\[1ex] 
\multirow{2}{*}{\textbf{Inc-6}} & \textbf{GeppNet}     & 93.1 & 93.0 & 92.9 \\
                                & \textbf{GeppNet+STM} & 92.7 & 89.9 & 91.4 \\[1ex] 
\multirow{2}{*}{\textbf{Inc-7}} & \textbf{GeppNet}     & 93.7 & 83.9 & 92.2 \\
                                & \textbf{GeppNet+STM} & 92.7 & 77.4 & 85.4 \\[1ex] 
\multirow{2}{*}{\textbf{Inc-8}} & \textbf{GeppNet}     & 94.5 & 73.6 & 92.1 \\
                                & \textbf{GeppNet+STM} & 94.0 & 74.7 & 90.4 \\[1ex] 
\multirow{2}{*}{\textbf{Inc-9}} & \textbf{GeppNet}     & 94.8 & 74.0 & 91.0 \\
                                & \textbf{GeppNet+STM} & 94.6 & 55.0 & 89.6 \\ 
\bottomrule \\
 
\end{tabular}
\caption{Tests to verify that GeppNet and GeppNet+STM were correctly implemented.  
These tests use the parameters and training strategy from \inline{Gepperth2016}.}
\label{table:som_verify}
\end{table}

Table \ref{table:fel_verify} shows the results from the FEL verification experiment.  We reproduced the non-stationary MNIST classification task with all ten digits proposed by \inline{FEL}.  Complete reproducibility was difficult because the authors do not state the learning rate or number of epochs for training, but the results are still comparable.

\begin{table}[ht!]
\centering \footnotesize
\begin{tabular}{@{}cc@{}}\toprule
\textbf{Non-Stationary} & \textbf{FEL Network}  \\
\textbf{Percentage} & \textbf{Performance} \\
\midrule
0.00 & 86.2 \\ 
0.25 & 67.9 \\
0.50 & 55.2 \\
0.75 & 46.7 \\
1.00 & 46.2 \\
\bottomrule \\

\end{tabular}
\caption{Tests to verify that FEL was correctly implemented.  Tests match parameters and training strategy from \inline{FEL}.}
\label{table:fel_verify}
\end{table}

\subsection{Plots and Tables for Experimental Results}

In this section we provide plots and tables demonstrating the performance of the various models on the data permutation, incremental class learning, and multi-modal experiments. Additionally, we provide a comparison of results on the MNIST dataset to results on the harder CUB-200 and AudioSet datasets.

\subsubsection{Data Permutation Experiment}

\begin{figure*}[ht!]
    \centering
    \subfigure[First Permutation - MNIST]{
        \centering
        \includegraphics[width=0.45\linewidth]{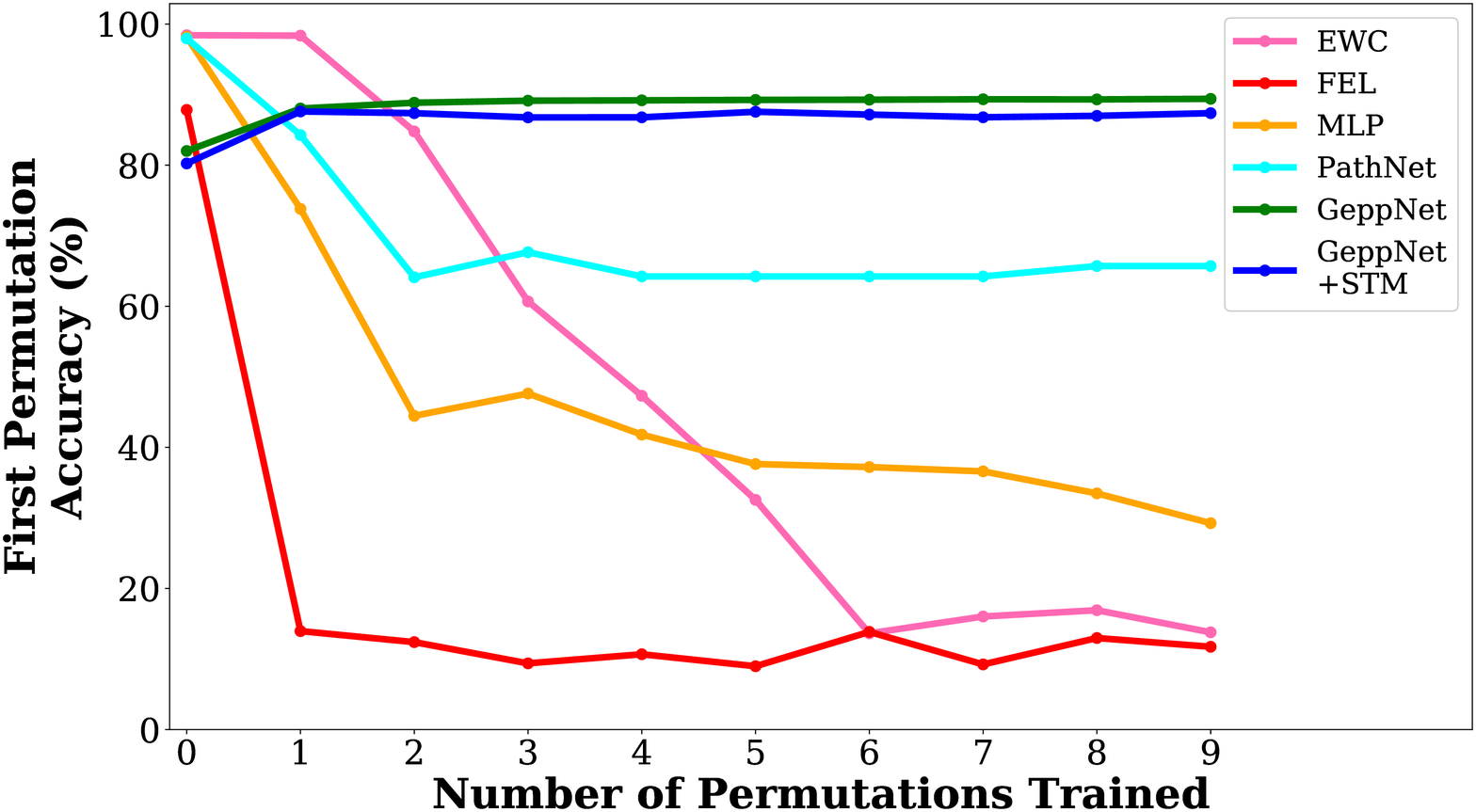}
        \label{fig:multitask-mnista}
}
     \subfigure[Most Recent Permutation - MNIST]{
        \centering
        \includegraphics[width=0.45\linewidth]{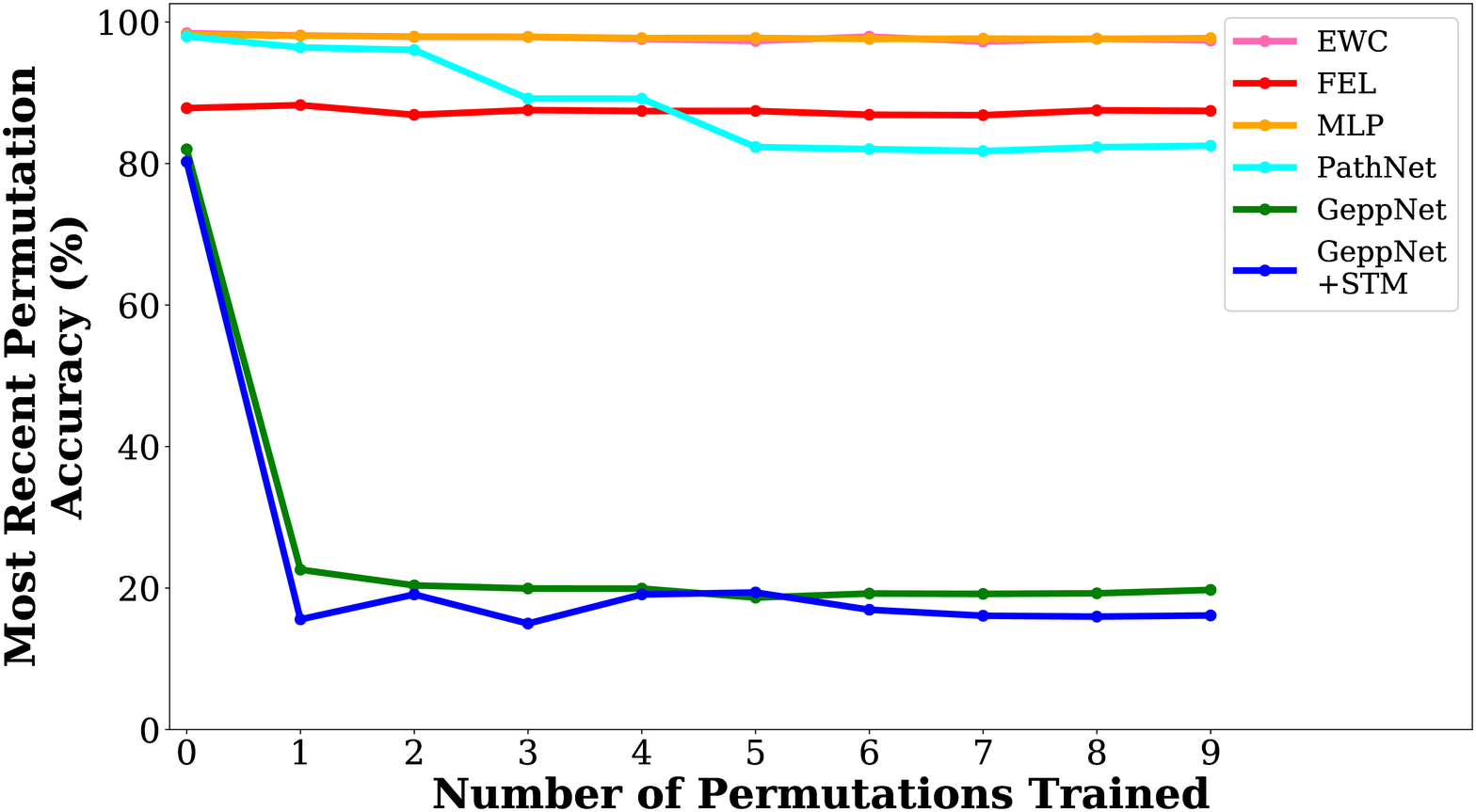}
        \label{fig:multitask-mnistb}}
    \subfigure[First Permutation - CUB-200]{
        \centering
        \includegraphics[width=0.45\linewidth]{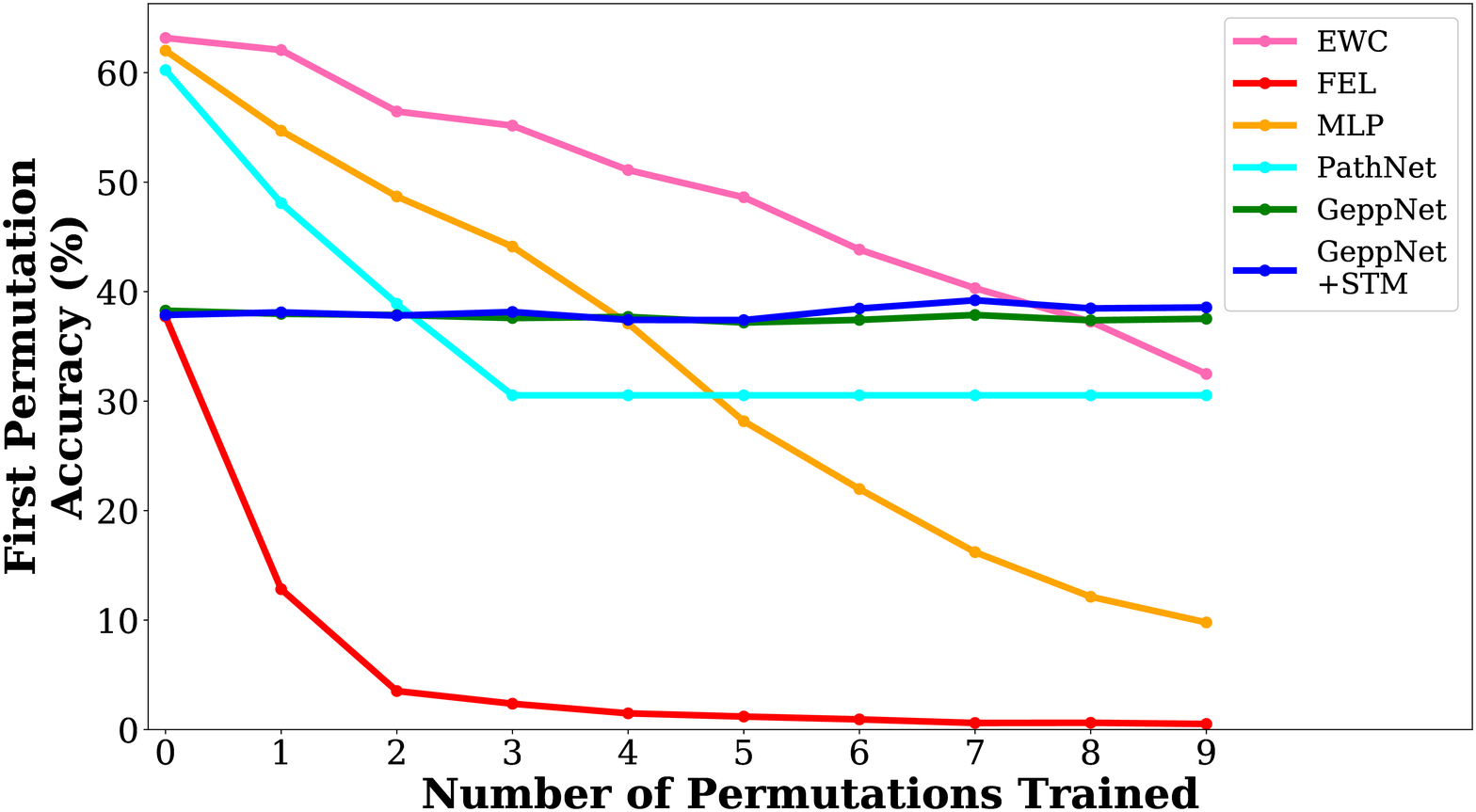}
        \label{fig:multitask-cuba}
}
    \subfigure[Most Recent Permutation - CUB-200]{
        \centering
        \includegraphics[width=0.45\linewidth]{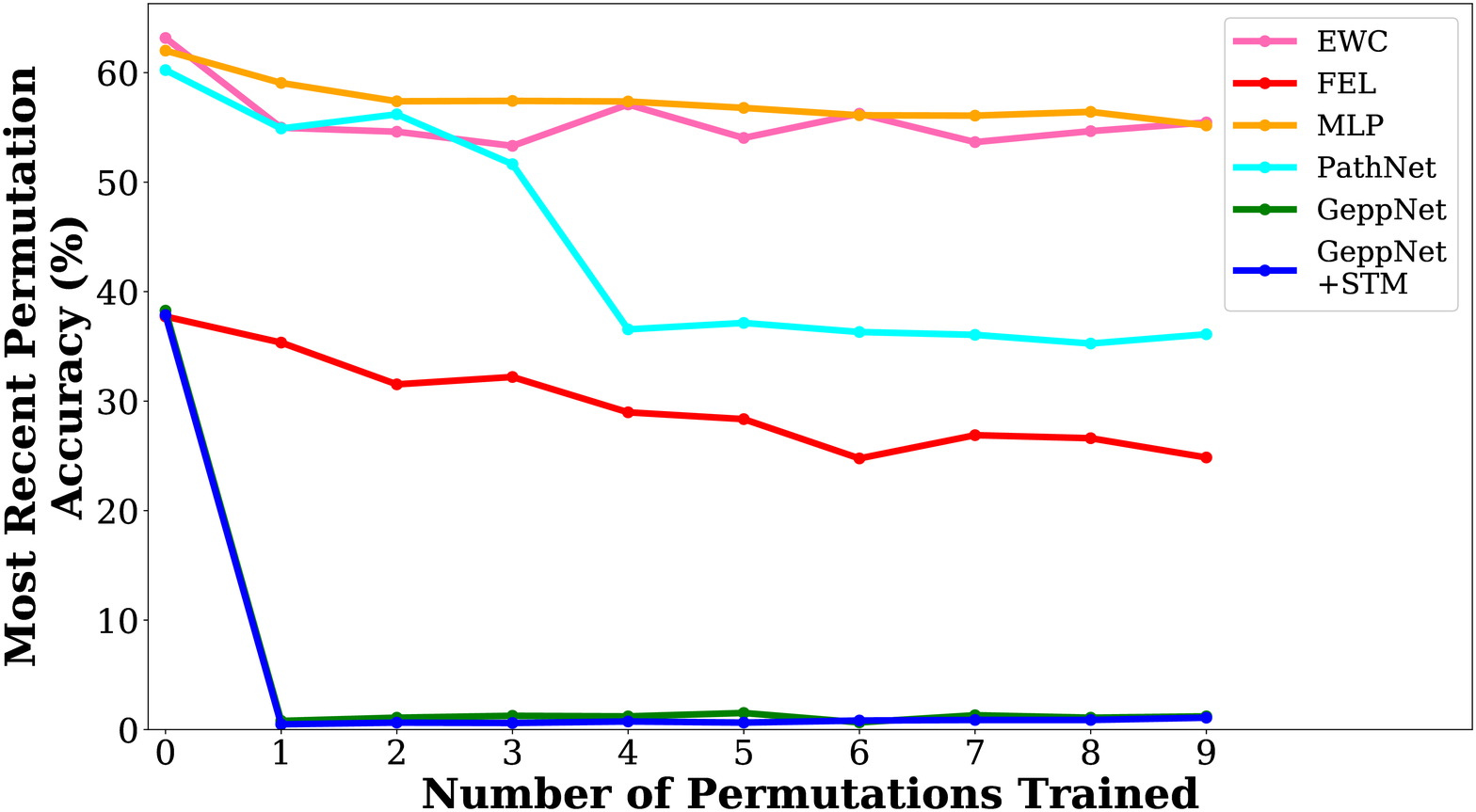}
        \label{fig:multitask-cubb}
        }
    \subfigure[First Permutation - AudioSet]{
        \centering
        \includegraphics[width=0.45\linewidth]{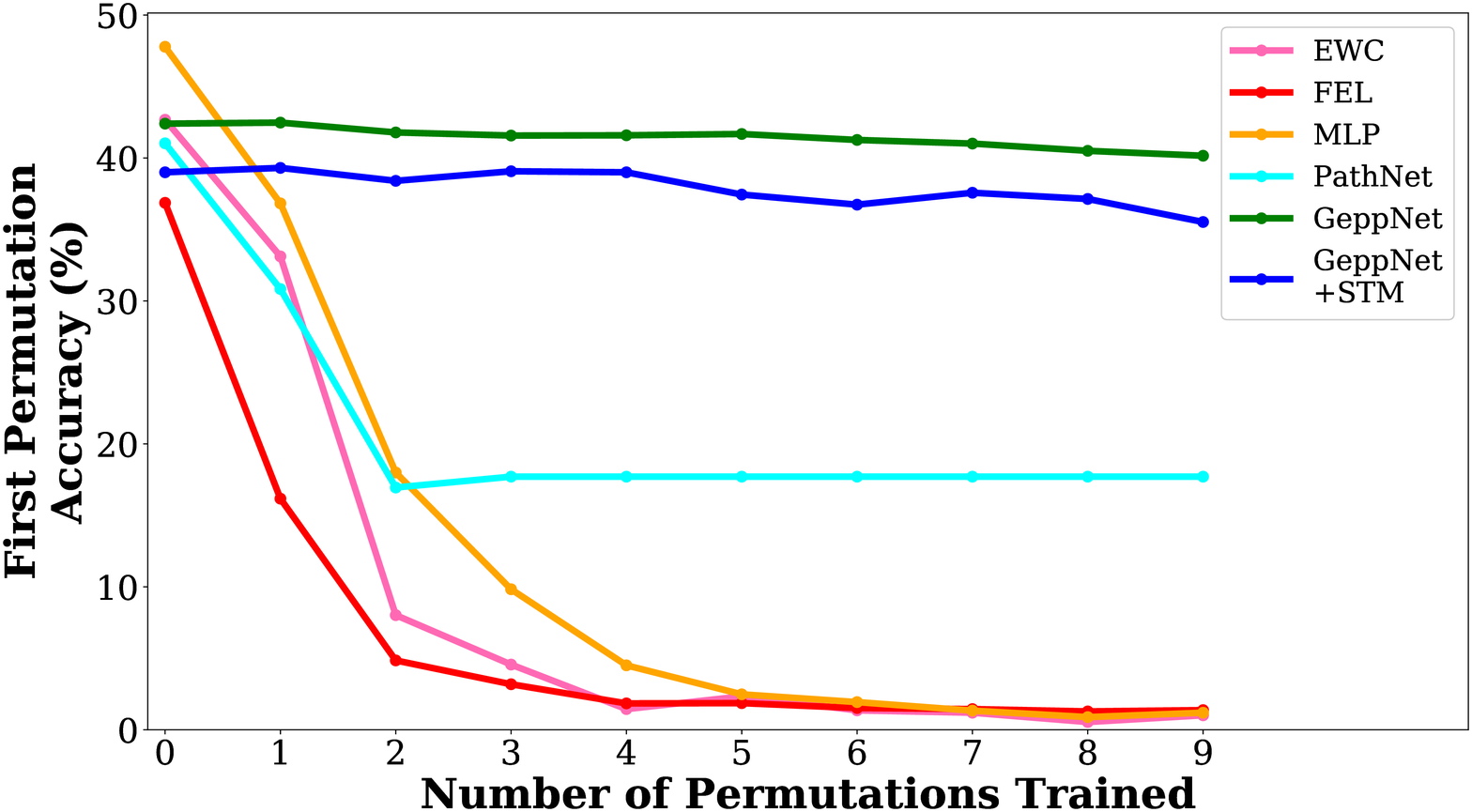}
        \label{fig:multitask-asa}}
    \label{fig:multitask-as}
    \subfigure[Most Recent Permutation - AudioSet]{
        \centering
        \includegraphics[width=0.45\linewidth]{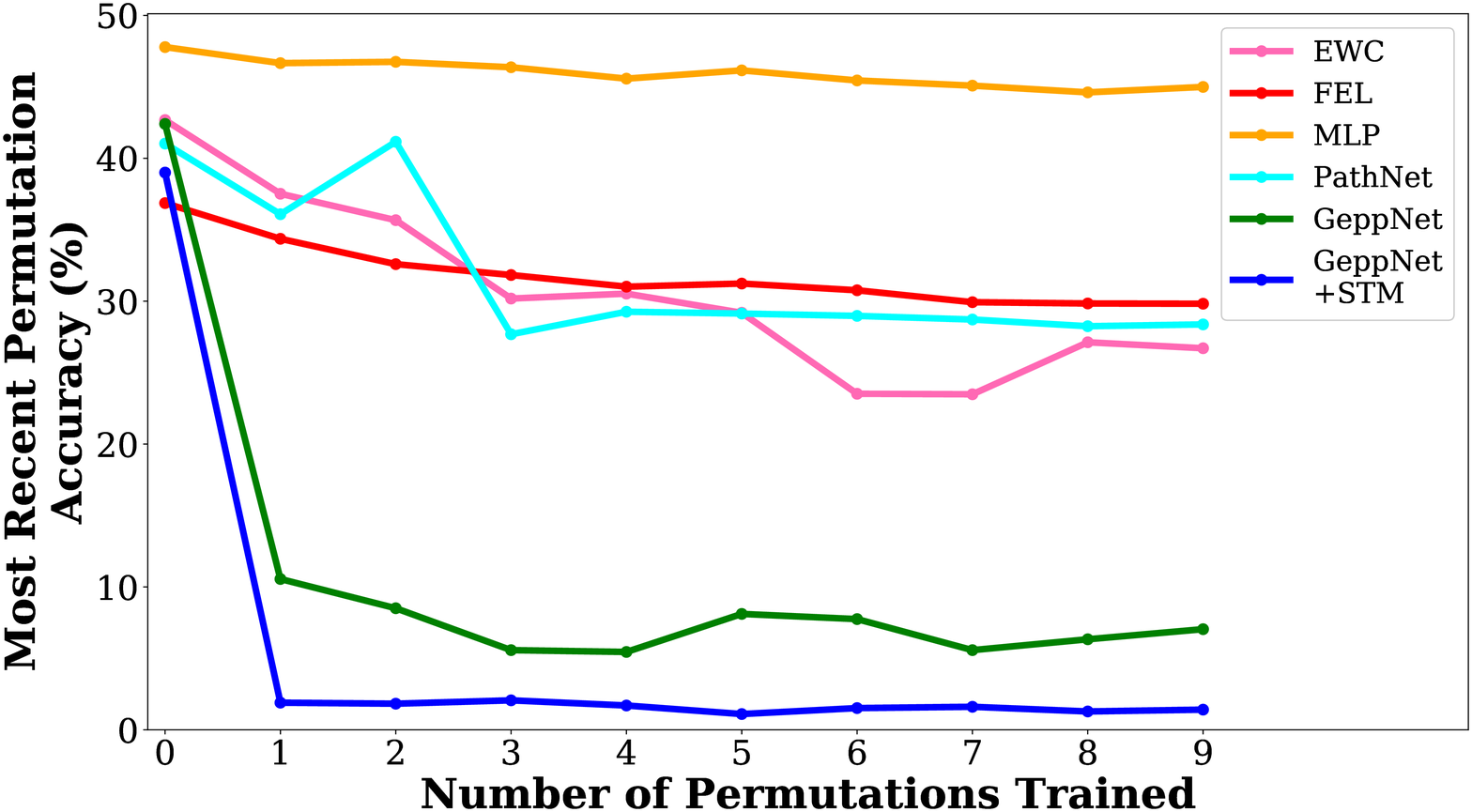}
        \label{fig:multitask-asb}
		}
    
    \caption{Data Permutation experiment for MNIST, CUB-200, and AudioSet.  The first column shows the performance of the original task as new tasks are learned and the second column shows the performance of the most recent permutation.} 
    \label{fig:multitask}
\end{figure*}

Fig. \ref{fig:multitask} shows the results of the data permutation experiment on the MNIST, CUB-200, and AudioSet datasets.  The first column of Fig. \ref{fig:multitask} shows the performance of the first session (original data) as the network learns new permutations and the second column of Fig. \ref{fig:multitask} shows the performance of the current permutation to demonstrate that the network is still learning new information.  Although GeppNet and GeppNet+STM appear to be retaining the original task, they do not appear to be acquiring new information. 

While performance is worse for all models on the CUB-200 and AudioSet datasets than on MNIST (See Fig. \ref{fig:multitask} and Table \ref{table:results}), some models exhibit similar trends in behavior independent of the dataset. In particular, GeppNet and GeppNet+STM retain the original data, but are unable to learn new information for both the CUB-200 and AudioSet datasets, which is similar to the behavior they exhibited on MNIST. In addition, FEL is prone to catastrophically forgetting the original data while maintaining the ability to learn new information, with worse performance than the MLP for learning new information on all three datasets.

While EWC and PathNet have the best overall performance, both models perform  worse on the CUB-200 and AudioSet datasets than on MNIST. Although PathNet is able to retain some knowledge of the original data while still maintaining the ability to learn new information on the CUB-200 and AudioSet datasets, its retention accuracy and newly trained task accuracy are much lower than in the MNIST experiments. Additionally, the EWC and MLP models exhibit similar behavior to one another on AudioSet with both models catastrophically forgetting the original data, while still maintaining some ability to learn new information. 

For the permutation task on the CUB-200 dataset, EWC performs the best, with similar trends to its performance on MNIST. With many of the models yielding significantly different trends and worse overall performance for the permutation task on the CUB-200 and AudioSet, it is important to consider scalability to large datasets before choosing a model for an incremental learning based task.

\subsubsection{Incremental Class Learning Experiment}

\begin{figure*}[th!]
  \centering
      \subfigure[Base Set Accuracy for MNIST]{
        \includegraphics[width=0.45\linewidth]{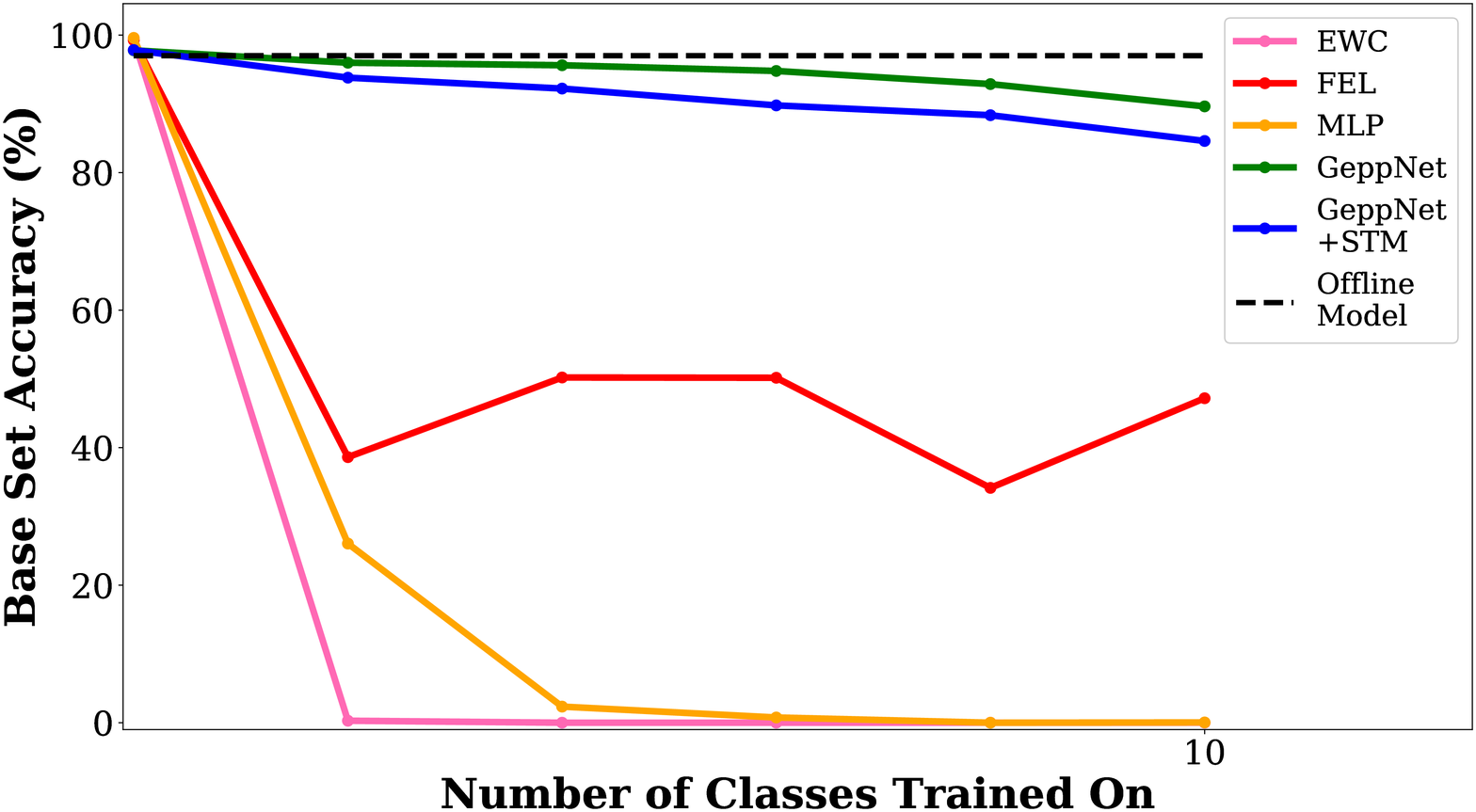}
        \label{fig:incremental-mnista}}
    \subfigure[Overall Accuracy for MNIST]{
        \includegraphics[width=0.45\linewidth]{figures/inc_all_mnist.eps}
  	\label{fig:incremental-mnistb}}
  \subfigure[Base Set Accuracy for CUB-200]{
  \includegraphics[width=0.45\linewidth]{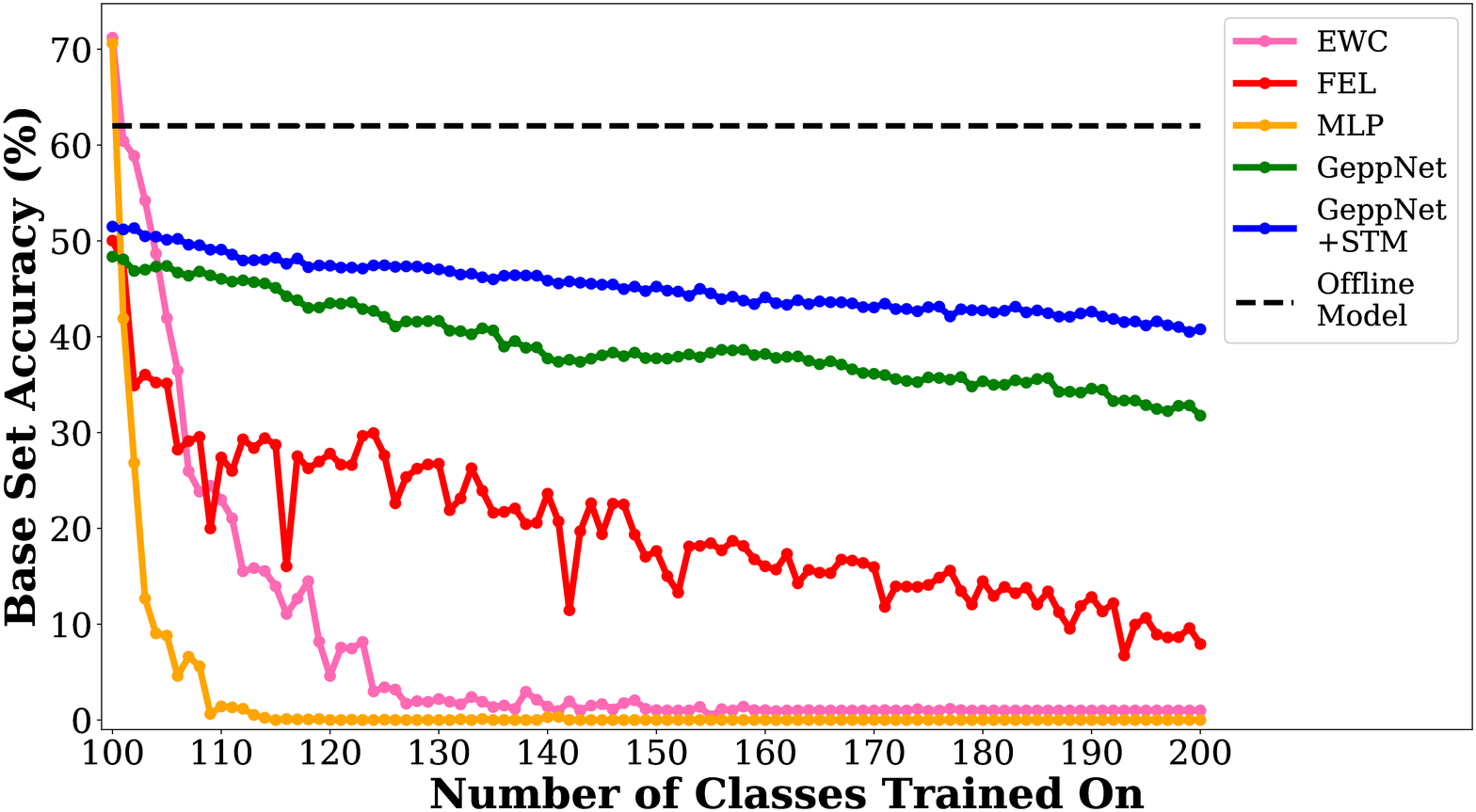}
  \label{fig:base_cub}}
  \subfigure[Overall Accuracy for CUB-200]{
  \includegraphics[width=0.45\linewidth]{figures/inc_all_cub.eps}
  \label{fig:overall_cub}}  
  \subfigure[Base Set Accuracy for AudioSet]{
  \includegraphics[width=0.45\linewidth]{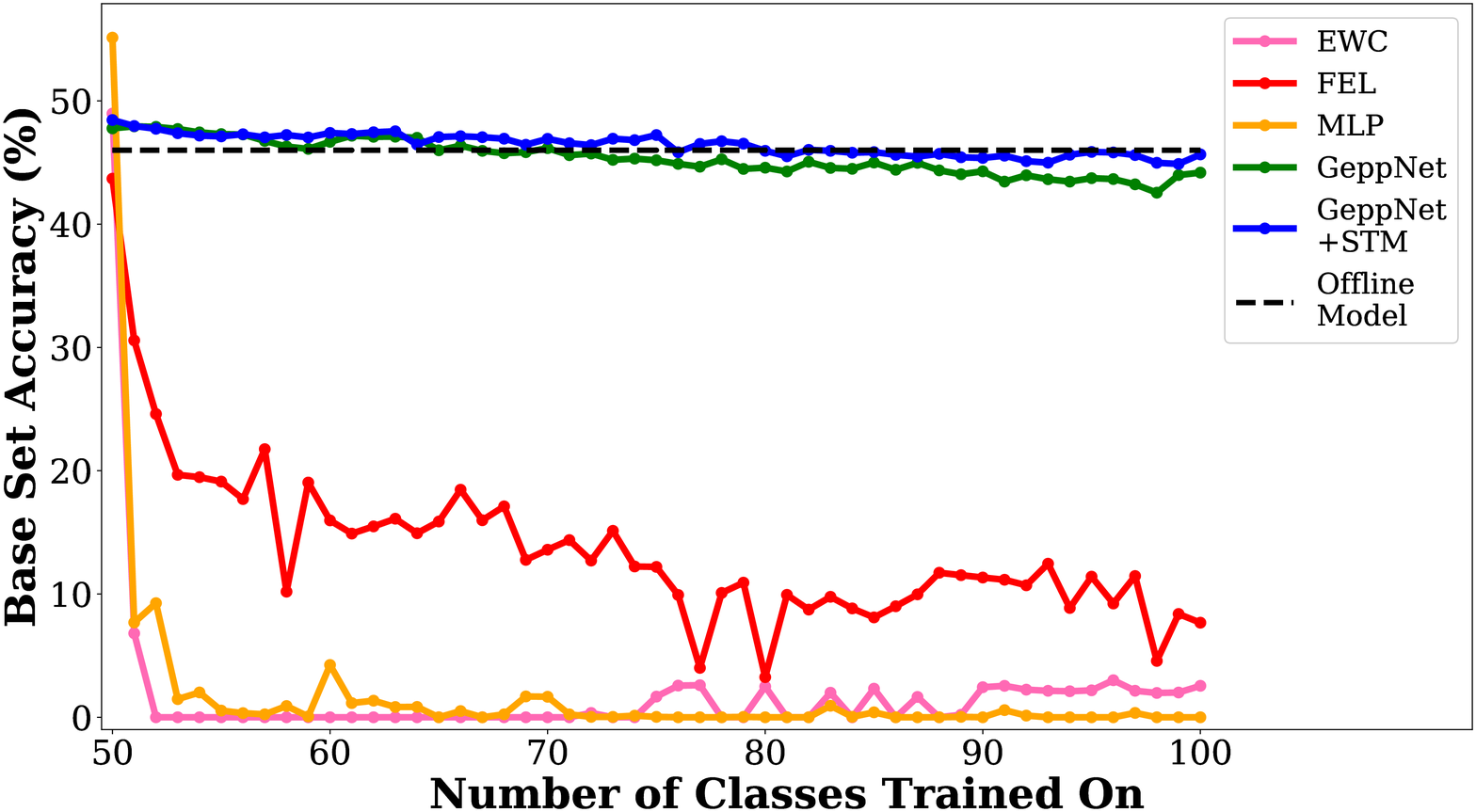}
  \label{fig:base_as}}
  \subfigure[Overall Accuracy for AudioSet]{
  \includegraphics[width=0.45\linewidth]{figures/inc_all_as.eps}
  \label{fig:overall_as}}

  \caption{The results from the incremental learning experiment for MNIST, CUB-200, and AudioSet. The results are the mean-per-class accuracy of each model for the entire testing set over time. The first column shows the base set accuracy and the second column shows the overall accuracy. The dashed line shows the performance of the offline MLP. }
  \label{fig:incremental}
\end{figure*}

Fig. \ref{fig:incremental} shows the results of the incremental class learning experiment.  Similar to the permutation task, results for the incremental class learning experiment (See Fig. \ref{fig:incremental} and Table \ref{table:results}) on the CUB-200 and AudioSet are much worse than on MNIST. Overall, MLP and EWC do not perform well for the incremental task and GeppNet, GeppNet+STM, and FEL perform the best on all three datasets, with significantly better results on the MNIST dataset.  Both GeppNet and GeppNet+STM are capable of retaining prior knowledge while also learning new classes; however, GeppNet performs better since it trains for every iteration (instead of only during the sleep phase).   

\subsubsection{Multi-Modal Experiment}

Table \ref{table:multimodal2} shows the results for the multi-modal experiment. The results indicate that EWC performs the best for this task, which is consistent with the results presented in Table \ref{table:results}.

\begin{table}[t!]
\centering \footnotesize
\begin{tabular}{@{}cccc@{}}\toprule
 &  & \multicolumn{2}{c}{\textbf{Accuracy}} \\
 &  & \textbf{Initial} & \textbf{Final}   \\
\midrule
 \multirow{2}{*}{\textbf{MLP}}	 &\textbf{CUB/AS} & 61.8 / 41.2 & 20.3 / 41.2  \\
 				                  &\textbf{AS/CUB} & 47.7 / 60.9 & 9.1 / 60.9   \\[1ex]
 \multirow{2}{*}{\textbf{EWC}}	  &\textbf{CUB/AS} & 64.3 / 36.9 & 58.6 / 36.9  \\
 				                   &\textbf{AS/CUB} & 47.4 / 58.8 & 47.1 / 58.8  \\[1ex]
 \multirow{2}{*}{\textbf{PathNet}} &\textbf{CUB/AS} & 59.2 / 37.6 & 56.4 / 37.6  \\
 				                 &\textbf{AS/CUB} & 44.7 / 54.0 & 3.2 / 54.0  \\[1ex]
 \multirow{2}{*}{\textbf{GeppNet}}	 &\textbf{CUB/AS} & 38.3 / 1.0 & 9.7 / 1.0 \\
 				                 &\textbf{AS/CUB} & 41.8 / 0.5 & 42.1/ 0.5  \\[1ex]
 \multirow{2}{*}{\textbf{GeppNet+STM}}	 				 &\textbf{CUB/AS} & 36.9 / 1.0 & 1.9 / 1.0 \\
 							     &\textbf{AS/CUB} & 40.1 / 0.5 & 38.2 / 0.5  \\[1ex]
 \multirow{2}{*}{\textbf{FEL}}	 &\textbf{CUB/AS} & 40.5 / 32.9 & 6.8 / 32.9 \\
 				                 &\textbf{AS/CUB} & 33.8 / 32.0 & 21.8 / 32.0  \\
\bottomrule 
\end{tabular}
\caption{\textit{Results from Multi-Modal Experiment.} AS denotes AudioSet. For each experiment, ``A/B'' indicates where task A is trained first followed by the training of task B. Initial Accuracy is the performance for tasks A/B immediately after each are trained. Final Accuracy is the performance for each task after both tasks are trained.}
\label{table:multimodal2}
\end{table}

\subsection{Fast Correlation Based Filter}

In this paper, we used the Fast Correlation Based Filter (FCBF) to measure feature redundancy in each dataset~\cite{yu2003feature}.  FCBF uses symmetric uncertainty to measure the independence (inverse redundancy) between two random variables $X,Y$.  Symmetric uncertainty is defined in Eq. \ref{eq:symmetric_uncertainty} where $H\left(X\right)$ is the entropy of $X$ (Eq. \ref{eq:entropy}), $H\left(X\vert Y \right)$ is the entropy of $X$ after observing $Y$ (Eq. \ref{eq:conditional}), and $IG\left(X\vert Y\right)$ is the information gain between $X$ and $Y$ (Eq. \ref{eq:information_gain}).  

\begin{equation}
\label{eq:symmetric_uncertainty}
SU\left(X,Y\right) = 2\cdot\frac{IG\left(X\vert Y\right)}{H\left(X\right) + H\left(Y\right)}
\end{equation}

\begin{equation}
\label{eq:entropy}
H\left(X\right) = - \sum_i P\left(x_i\right) log_2\left(P\left(x_i\right)\right)
\end{equation}

\begin{equation}
\label{eq:conditional}
H\left(X \vert Y\right) = - \sum_j P\left(y_j\right) \sum_i P\left(x_i \vert y_j\right) log_2\left(P\left(x_i \vert y_j\right)\right)
\end{equation}

\begin{equation}
\label{eq:information_gain}
IG\left(X\vert Y\right) = H\left(X\right) - H\left(X \vert Y\right)
\end{equation}

Table \ref{table:fcbf} shows the total number of non-redundant features for each dataset along with the percentage of features that are not redundant in each dataset.  The results show that the features in MNIST and AudioSet are noticeably more redundant than the features found in CUB-200.

\begin{table}[t!]
\centering \footnotesize
\begin{tabular}{@{}ccc@{}}\toprule
\multirow{2}{*}{\textbf{Dataset}} & \textbf{Non-Redundant} & \textbf{Percentage of}  \\
 & \textbf{Features} & \textbf{Total Features} \\
\midrule
MNIST & 39 & 5.0\% \\
AudioSet & 129 & 10.1\% \\
CUB-200 & 450 & 22.0\% \\
\bottomrule 
\end{tabular}
\caption{\textit{Non-redundant features in MNIST, AudioSet, and CUB-200 datasets.}  This was determined using the Fast Correlation Based Filter algorithm.}
\label{table:fcbf}
\end{table}

\begin{figure}[th!]
  \centering

  \subfigure[MNIST]{%
  \includegraphics[width=0.9\linewidth]{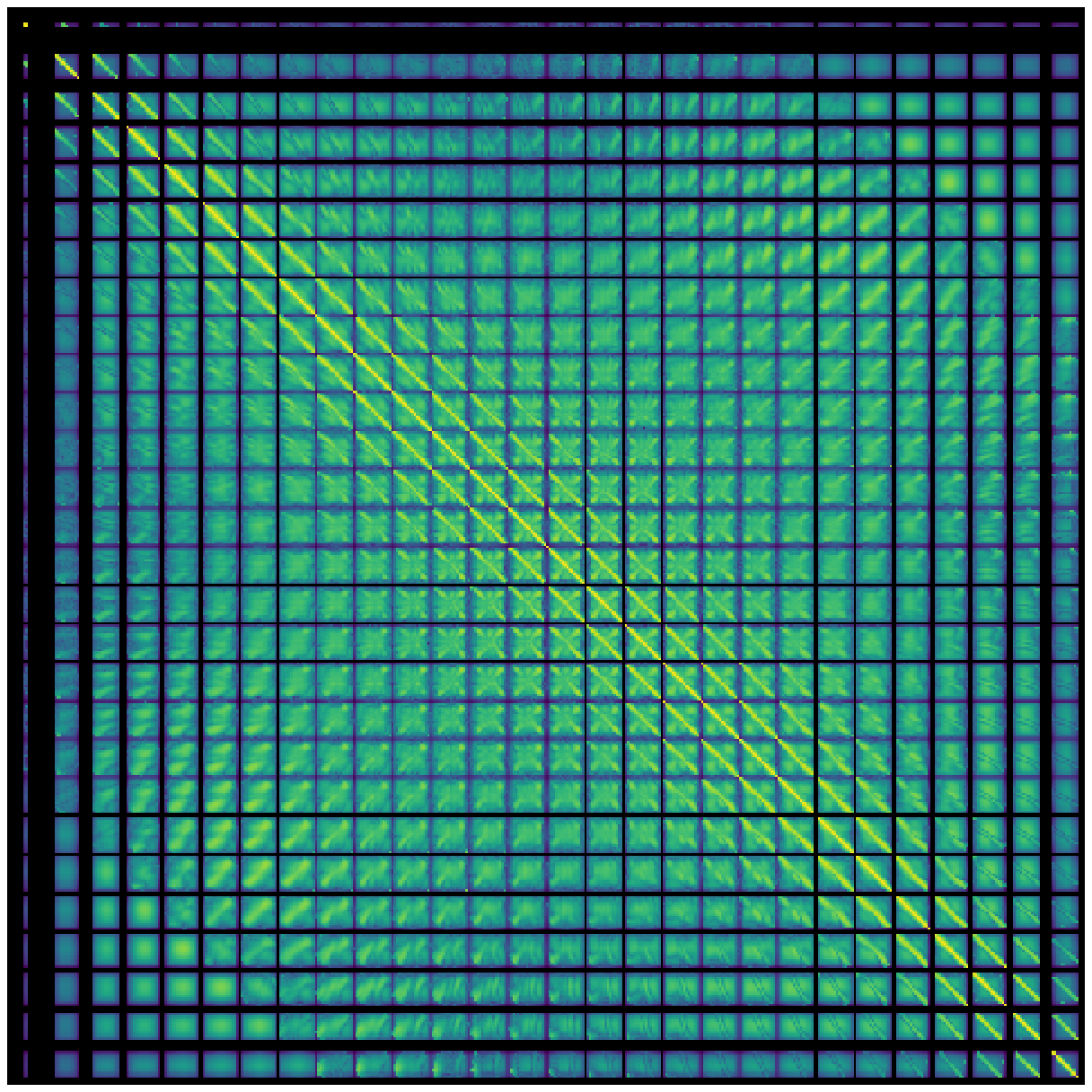}
  \label{fig:su_mnist}}
 
  \subfigure[CUB-200]{%
  \includegraphics[width=0.9\linewidth]{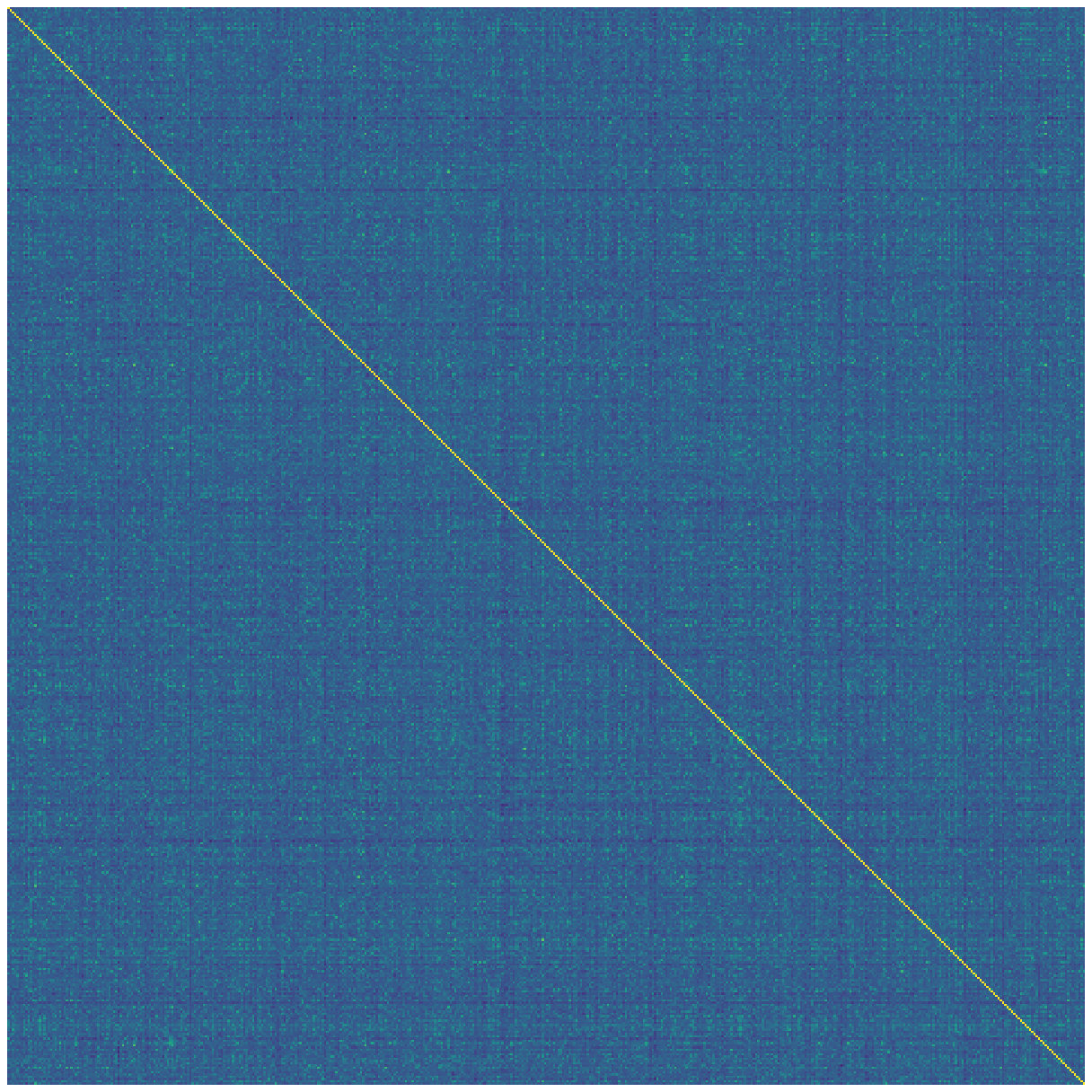}
  \label{fig:su_cub}} 
  
  \subfigure[AudioSet]{%
  \includegraphics[width=0.9\linewidth]{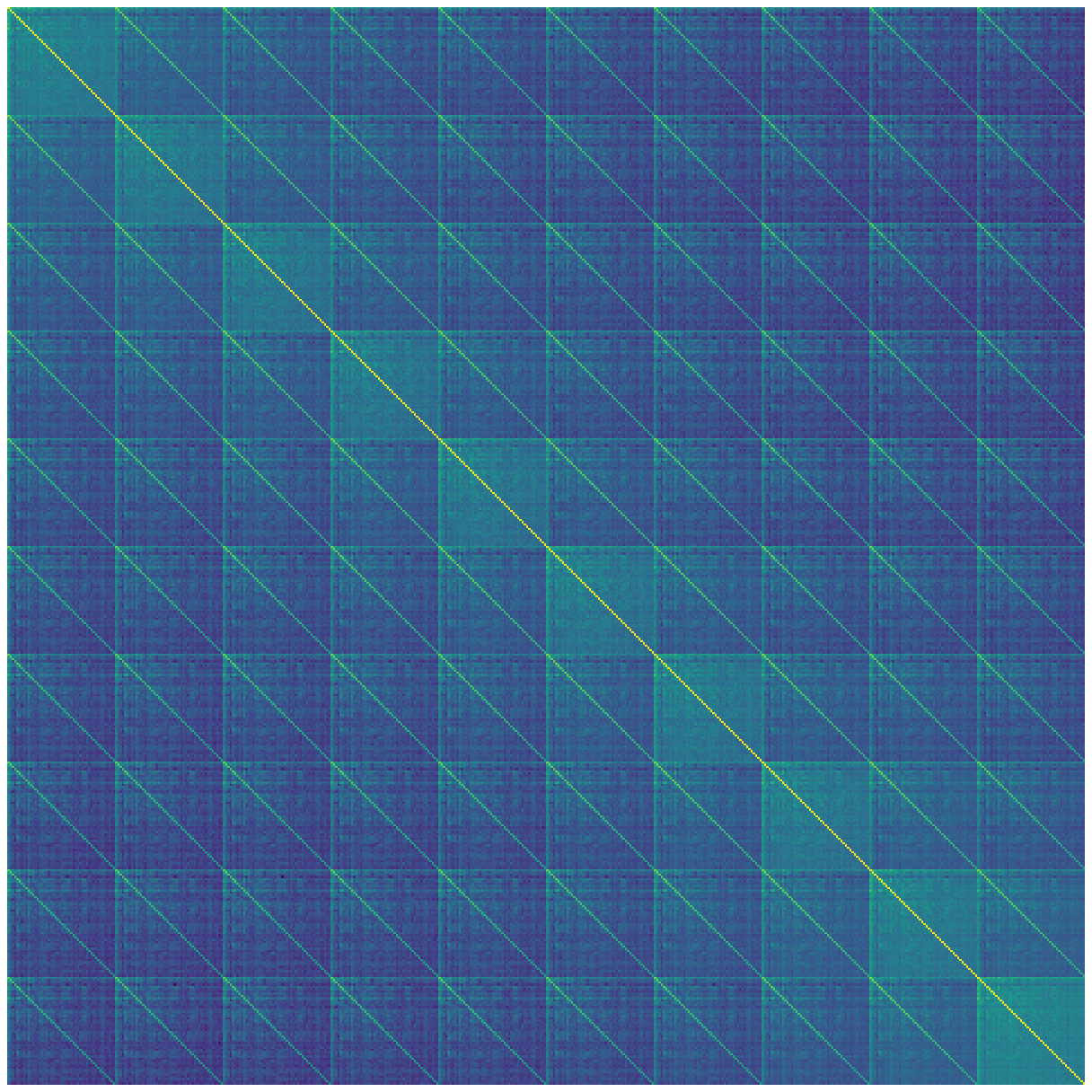}
  \label{fig:su_as}}

  \caption{\textit{Symmetric uncertainty coefficients for all three datasets.}}
  \label{fig:fcbf}
\end{figure}

Figure \ref{fig:fcbf} is a visualization of these results, where we show the symmetric uncertainty matrix for each dataset.  The results are a $F\times F$ matrix where $F$ is the dimensionality of the feature vector (e.g. CUB-200 is 2048).  The bright areas represent features that are strongly correlated with one another (they are more redundant).  The results show significant feature overlap in MNIST which is expected since it is a gray-scale image with zero values in the background.  AudioSet also has sub-diagonals across the matrix which correspond to highly correlated features that repeat over some interval.  Each AudioSet sample consists of ten 128-dimensional sub-vectors that are concatenated together to form a single vector.  Each sub-vector is the feature representation of the audio signal for a single second.  Since the sounds repeat across the entire ten seconds, the corresponding features are strongly correlated in those locations.  The CUB-200 features appear to not be strongly correlated.  This is probably because each sample is the ResNet-50 feature representation which is highly discriminative.

\subsection{Ideal Model}

Table \ref{table:results_ideal} show the experimental results when the model capacity is not constrained; that is, we performed a hyperparameter search to find the best model for each model/dataset combination.  The base results are a bit higher than the results where we constrained the model capacity (Table \ref{table:results}), but the main conclusions remain the same.

\begin{table*}[t!]
\centering \footnotesize
\begin{tabular}{@{}cc|ccc|ccc|c|c@{}}\toprule
 \multirow{2}{*}{\textbf{Model}} & \multirow{2}{*}{\textbf{Dataset}}  & \multicolumn{3}{c|}{\textbf{Data Permutation}} & \multicolumn{3}{c|}{\textbf{Incremental Class}} & \textbf{Memory} & \textbf{Model}  \\
 & & $\Omega_{base}$ & $\Omega_{new}$ & $\Omega_{all}$& $\Omega_{base}$ & $\Omega_{new}$ & $\Omega_{all}$  & \textbf{Constraints} & \textbf{Size (MB)} \\
\midrule
 \multirow{2}{*}{\textbf{MLP}}	  &\textbf{CUB}   & 0.449 & 0.936 & 0.619 & 0.000 & 0.640 & 0.011 & \multirow{2}{*}{Fixed-size} & 36.54  \\
 				                  &\textbf{AS}    & 0.336 & 0.950 & 0.578 & 0.025 & 1.000 & 0.050 &  &  4.44 \\[1ex]
 \multirow{2}{*}{\textbf{EWC}}	  &\textbf{CUB}   & 0.426 & 0.830 & 0.525 & 0.362 & 0.010 & 0.302 & \multirow{2}{*}{Fixed-size} & 13.19\\
 				                  &\textbf{AS}    & 0.118 & 0.459 & 0.182 & 0.249 & 0.000 & 0.213 &  & 4.41 \\[1ex]
 \multirow{2}{*}{\textbf{PathNet}}&\textbf{CUB}   & 0.538 & 0.701 & 0.655 &  N/A  &  N/A  &  N/A  & New output layer & 7.46\\
 				                  &\textbf{AS}    & 0.414 & 0.750 & 0.615 &  N/A  &  N/A  &  N/A  & for each task & 4.68\\[1ex]
 \multirow{2}{*}{\textbf{GeppNet}}	&\textbf{CUB} & 0.571 & 0.112 & 0.167 & 0.758 & 0.558 & 0.675 & Stores all & 58.33 \\
 				                  &\textbf{AS}    & 0.877 & 0.238 & 0.346 & 1.024 & 0.495 & 0.972 & training data & 153.12 \\[1ex]
 \multirow{2}{*}{\textbf{GeppNet+STM}}&\textbf{CUB}& 0.610 & 0.014 & 0.137 & 0.803 & 0.217 & 0.686 & Stores all & 59.77 \\
							      &\textbf{AS}    & 0.857 & 0.125 & 0.272 & 1.025 & 0.372 & 0.942 & training data & 153.94 \\[1ex]
 \multirow{2}{*}{\textbf{FEL}}	  &\textbf{CUB}   & 0.575 & 0.880 & 0.732 & 0.735 & 0.976 & 0.672 & \multirow{2}{*}{Fixed-size} & 209.06 \\
 				                  &\textbf{AS}    & 0.191 & 0.853 & 0.444 & 0.595 & 0.999 & 0.541 & & 247.07 \\[1ex]
\bottomrule 
\end{tabular}
\caption{Results on CUB-200 (CUB) and AudioSet (AS) for our evaluation metrics as well as model size (in MB) for each model/dataset combination.  In these experiments, we optimized the model capacity and other hyperparameters for each model/dataset combination.}
\label{table:results_ideal}
\end{table*}

\end{document}